\documentclass[sigconf]{acmart}

\usepackage{balance}    %
\usepackage{color}
\usepackage{color, colortbl}
  
\usepackage{subcaption}
\usepackage{multirow}
\usepackage{siunitx}

\usepackage{xspace}

\usepackage{graphicx}
\usepackage{xcolor}
\usepackage{soul}

\AtBeginDocument{%
  \providecommand\BibTeX{{%
    \normalfont B\kern-0.5em{\scshape i\kern-0.25em b}\kern-0.8em\TeX}}}

\copyrightyear{2023}
\acmYear{2023}
\setcopyright{rightsretained}
\acmConference[CHI '23]{Proceedings of the 2023 CHI Conference on Human Factors in Computing Systems}{April 23--28, 2023}{Hamburg, Germany}
\acmBooktitle{Proceedings of the 2023 CHI Conference on Human Factors in Computing Systems (CHI '23), April 23--28, 2023, Hamburg, Germany}\acmDOI{10.1145/3544548.3581254}
\acmISBN{978-1-4503-9421-5/23/04}

\begin{document}

\title{Toucha11y: Making Inaccessible Public Touchscreens Accessible}

\author{Jiasheng Li}
\email{jsli@umd.edu}
\affiliation{%
  \institution{University of Maryland}
  \city{College Park}
  \state{Maryland}
  \country{USA}
}
\author{Zeyu Yan}
\email{zeyuy@umd.edu}
\affiliation{%
  \institution{University of Maryland}
  \city{College Park}
  \state{Maryland}
  \country{USA}
}
\author{Arush Shah}
\email{arushshah1@gmail.com}
\affiliation{%
  \institution{University of Maryland}
  \city{College Park}
  \state{Maryland}
  \country{USA}
}
\author{Jonathan Lazar}
\email{jlazar@umd.edu}
\affiliation{%
  \institution{University of Maryland}
  \city{College Park}
  \state{Maryland}
  \country{USA}
}
\author{Huaishu Peng}
\email{huaishu@umd.edu}
\affiliation{%
  \institution{University of Maryland}
  \city{College Park}
  \state{Maryland}
  \country{USA}
}

\renewcommand{\shortauthors}{Li et al.}

\begin{abstract}
Despite their growing popularity, many public kiosks with touchscreens are inaccessible to blind people.
Toucha11y is a working prototype that allows blind users to use existing inaccessible touchscreen kiosks independently and with little effort.
Toucha11y consists of a mechanical bot that can be instrumented to an arbitrary touchscreen kiosk by a blind user and a companion app on their smartphone. 
The bot, once attached to a touchscreen, will recognize its content, retrieve the corresponding information from a database, and render it on the user's smartphone. 
As a result, a blind person can use the smartphone's built-in accessibility features to access content and make selections.
The mechanical bot will detect and activate the corresponding touchscreen interface.
We present the system design of Toucha11y along with a series of technical evaluations.
Through a user study, we found out that Toucha11y could help blind users operate inaccessible touchscreen devices.
\end{abstract}

\begin{CCSXML}
<ccs2012>
   <concept>
       <concept_id>10003120.10011738.10011776</concept_id>
       <concept_desc>Human-centered computing~Accessibility systems and tools</concept_desc>
       <concept_significance>500</concept_significance>
       </concept>
 </ccs2012>
\end{CCSXML}

\ccsdesc[500]{Human-centered computing~Accessibility systems and tools}

\keywords{accessibility, touchscreen appliances, robotic, visual impairments}

\maketitle

\begin{figure}[ht!]
  \includegraphics[width=\columnwidth]{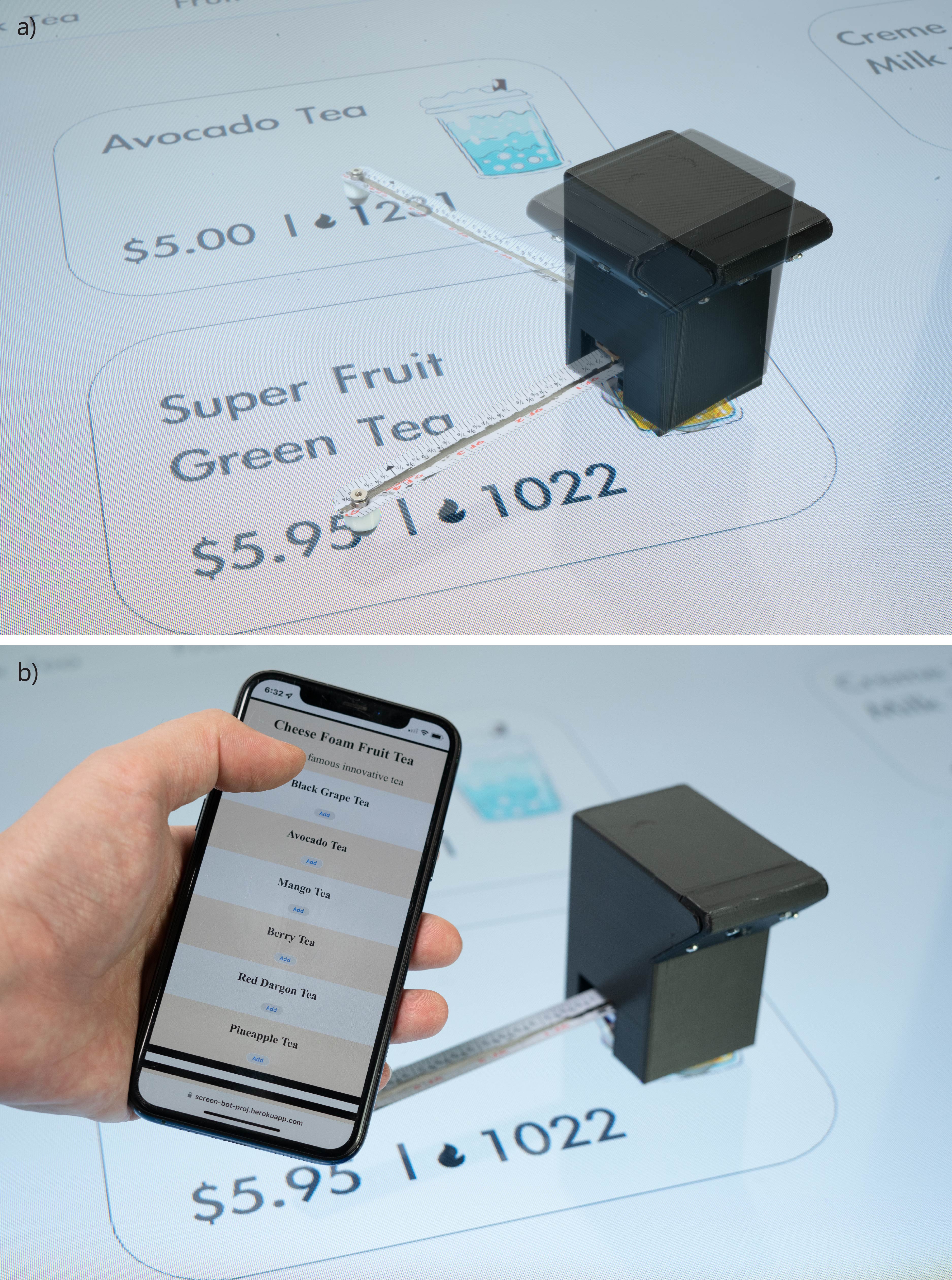}
  \caption{Toucha11y overview. a) The Toucha11y bot can be placed on top of an inaccessible touchscreen kiosk to activate its interface. b) A blind user can access and use the kiosk from their personal smartphone.}
  \Description{Toucha11y overview. a) The Toucha11y bot can be placed on top of an inaccessible touchscreen kiosk to activate its interface. b) A blind user can access and use the kiosk from their personal smartphone.}
  \label{fig:teaser}
\end{figure}

\section{Introduction}

From self-service food in restaurants to ID renewal in the Department of Motor Vehicles (DMV), touchscreen kiosks have been increasingly popular in workplaces and public access areas. 
The emergence of the COVID-19 pandemic further accelerates the adoption of touchscreen devices~\cite{covid}, as they provide customers with an autonomous experience while promoting social distancing. 
Unfortunately, the widespread use of touchscreen technology on kiosks can be challenging for blind people. 
Being primarily visual and nonhaptic, many touchscreen devices---especially those in public---are not accessible to them, preventing blind users from performing tasks independently and potentially introducing feelings of embarrassment, as discussed by Kane et al.~\cite{kane2009freedom}.

Much effort has recently been made to improve the accessibility of public touchscreen devices. 
Several recent lawsuits in the United States, for example, have compelled large corporations to make their public touchscreen devices accessible~\cite{legal_cases}. 
The Americans with Disabilities Act (ADA) has also required that public kiosks (such as ATMs and movie rental kiosks) be accessible to blind users~\cite{adaag}. 
Indeed, despite the slow pace, we are seeing an increasing number of public touchscreen devices outfitted with accessibility features~\cite{mcdonald}. 
Unfortunately, because the guidelines for designing accessible kiosks are not clearly defined, the majority of public touchscreens are still not useful to blind users~\cite{lazar2019toward}. 
Even devices with touchscreens and physical keypads, for example, may not have any audio output~\cite{lazar2019toward}; others with audio output may simply speak the information publicly, ignoring the privacy of blind users. 
Moreover, a significant number of inaccessible touchscreens have already been deployed worldwide. Waiting for device manufacturers to update all of them will not solve today's accessibility problems.

To make those already-implemented public devices accessible to blind users, researchers in the field of HCI have proposed different methods to augment or retrofit the touchscreens. 
For example, Flexible Access System~\cite{flexibleAccess}, Touchplates~\cite{kane2013touchplates}, and Facade~\cite{guo2017facade} propose enhancing existing touchscreens with (3D-printable) physical buttons and custom tactile overlays. 
Although promising, it is impractical for blind users to install custom guidance on arbitrary public touchscreens they encounter in their daily lives. 
VizLens~\cite{VizLens} proposes a crowdsourcing solution in which photos of a touchscreen interface taken by blind users are first labeled by crowd workers and then used to guide blind users to operate the touchscreen. 
This method works with static touchscreens (such as those on microwave ovens), but it can be difficult with touchscreens with changeable content.
To work with dynamic touchscreen interfaces, Statelens ~\cite{guo2019statelens} proposes reverse engineering their underlying state diagrams. 
It then generates instructions for blind users to use the touchscreens in conjunction with a custom 3D-printed accessory. 
Statelens allows blind users to explore arbitrary public touchscreens independently, but in order for the step-wise guidance to work, blind users must hold their phone with one hand and keep its camera focused on the touchscreen while exploring the screen interface with the other.
In such cases, bimanual operations can be difficult for some blind users.
Besides, since the content and layout of touchscreen interfaces can vary, physically learning and exploring a touchscreen interface can be time-consuming and laborious.

Building upon previous research, we present Toucha11y\footnote{Toucha11y is a combination of the words "touch" and "accessibility." The "11" refers to the 11 letters in "accessibility" between "a" and "y." See more at \url{https://www.a11yproject.com/}} , a working prototype that aims to enable blind users to access arbitrary public touchscreen devices independently and with little effort. The key to Toucha11y is to bridge the gap between blind users and a public touchscreen device with a set of hardware and software tools, allowing them to explore touchscreen content from their familiar smartphone devices without having to deal with the unfamiliar, inaccessible public touchscreen directly. To make the bridge work, Toucha11y's hardware---a small mechanical bot---must be placed on top of a public touchscreen device by a blind user (Figure~\ref{fig:teaser}a). Once placed, the bot's onboard camera will photograph the screen, with its corresponding interface (which can be generated through crowdsourcing and reverse engineering~\cite{guo2019statelens}) sent to the user's smartphone. The blind user can freely explore and select contents using the smartphone's built-in accessibility features such as Apple's VoiceOver and Android's TalkBack~\cite{apple_voiceover,android_talkback} (Figure~\ref{fig:teaser}b). These selections will be sent back to the bot, which will physically register the corresponding touch event for the blind user using an extendable reel. With Toucha11y, a blind user only needs to interact with an inaccessible touchscreen once, when placing the bot on the screen. Because the remainder of the interactions occurs on the user's smartphone, a blind user is not required to learn how to use a new touchscreen interface every time a new device is encountered. Toucha11y can also help alleviate privacy concerns by allowing blind users to use their personal devices for input and voice output, which they likely already have configured for privacy (with earphones or other options). 

Toucha11y's design is inspired by a list of previous works as well as an interview with nine blind participants. In the following sections, we will first present the interview findings and the distilled design guidance. We will then detail its system implementation, including the key features of the mechanical bot, the smartphone interface, and the companion back-end server. Through a series of technical evaluations, we demonstrate that the Toucha11y bot can accurately recognize the content of a touchscreen and its relative location. The extendable reel can also reach target locations with high accuracy. A user study with seven blind participants shows that Toucha11y can provide blind users with a solution to independently use an inaccessible touchscreen kiosk interface. 

In summary, our paper contributes to: 1) the investigation of blind users' practices and challenges in using public touchscreen devices; 2) Toucha11y, a working prototype that makes inaccessible touchscreen devices accessible to blind users; and 3) a user study to evaluate the tool.

\section{Related Work}
Our work builds upon the notions of 1) touchscreen accessibility, 2) personal assistive devices, and 3) accessibility with computer vision.

\subsection{Touchscreen Accessibility}
Over the last two decades, the widespread use of touchscreens on mobile phones and tablets has raised serious concerns about their accessibility for blind and low-vision users. In response, numerous efforts have been made in both academia and industry to increase their accessibility. For example, Talking Fingertip Technique~\cite{vanderheiden1996use}, Slide Rule~\cite{kane2008slide}, among others~\cite{yfantidis2006adaptive, 10.1145/1868914.1868972, 10.1145/1240624.1240836}, propose using (multi-)touch gestures to control touchscreens non-visually, which influenced the design of VoiceOver~\cite{apple_voiceover} and TalkBack~\cite{android_talkback}, that are now widely adopted in modern smartphone devices. Reports have shown that smartphones are now frequently used among blind users and are even replacing traditional solutions ~\cite{martiniello2022exploring}.

Touchscreens are also prevalent outside of personal computing devices today. Unfortunately, the growing number of public kiosks with touchscreens, such as those in hospitals, airports, markets, and restaurants, are not as accessible as smartphones. Despite efforts to improve their accessibility (e.g., ADA laws in the United States~\cite{adaag}), a large proportion of existing kiosks with touchscreens are inaccessible to blind people~\cite{lazar2019toward}. In order to prevent their exclusion, researchers have proposed techniques to make existing public touchscreen devices accessible. One approach is to provide haptics, physical buttons, or tactile displays to existing touchscreens~\cite{flexibleAccess, kane2013touchplates, guo2017facade, VizLens}. While additional haptics can help blind users understand a touchscreen interface, it is impractical for blind people to install bespoke assistance on random public touchscreens for touch guidance. Rather than retrofitting the touchscreen, Statelens~\cite{guo2019statelens} proposes to use computer vision and crowdsourcing to reverse engineer the underlying state diagrams of a touchscreen interface. From these state diagrams, step-by-step instructions can be generated to assist blind individuals in using touchscreens with a 3D-printed accessory.

Building upon the reverse engineering approach, Toucha11y assumes that labeled touchscreen interfaces can be retrieved from a shared repository or database. Unlike Statelens, Toucha11y moves the interface exploration from the inaccessible touchscreen to the users' phones. This way, blind users don't have to learn how to use each new public kiosk they encounter.

\subsection{Personal Assistive Devices}

While not widely used yet~\cite{milallos2021would, milallos2022exploratory}, personal assistive devices~\cite{velazquez2010wearable, ye2014current} have the potential to improve real-world accessibility for blind users in a variety of contexts, from providing travel and navigation aids ~\cite{velazquez2010wearable, dakopoulos2009wearable, elmannai2017sensor, wahab2011smart} to promoting individual creativity~\cite{10.1145/3526113.3545627, koushik2019storyblocks, davis2020tangiblecircuits}. Personal assistive devices have also been utilized to support situations such as shopping in order to promote independence. Shoptalk~\cite{nicholson2009shoptalk}, for instance, aims to improve the shopping experience of blind users with handheld barcode scanners. Foresee~\cite{zhao2015foresee} demonstrates how a head-mounted vision solution can magnify real-world objects to assist low-vision users in the grocery store.

Because of their embodiment possibilities and potential social presence, robots in various forms have been investigated as a means of assistive technology in addition to wearables or handheld devices. Mobile robots, like wearable devices, have been used for navigation and road guidance~\cite{kulyukin2004rfid, kulyukin2006robot, kulyukin2005robocart}. Drones have also been investigated for use in guiding blind runners and assisting with indoor navigation~\cite{al2016exploring, avila2017dronenavigator}.

Toucha11y is also a personal assistive robot in that it is designed to be carried by the blind user. Unlike the previously mentioned research focusing on navigation and travel aids, Toucha11y proposes to automate existing and inaccessible touchscreen activation for blind users, which, to the best of our knowledge, has not previously been investigated.

\subsection{Accessibility with Computer Vision}
A large number of assistive technologies are based on computer vision. For example, using optical character recognition (OCR), several systems (such as the KNFB Reader~\cite{knfb}) have been developed to assist blind people in reading visual text. Camera-based solutions, such as those attached to a table~\cite{kane2013access}, worn on the finger~\cite{stearns2016evaluating, nanayakkara2013eyering}, or held in the hand~\cite{clearspeech}, are proposed to recognize text in physical documents and allow a blind person to hear and interact with them. 

Recent advancements in deep learning have enabled commercial solutions such as Seeing AI~\cite{seeingai} and Aipoly~\cite{aipoly} to apply general object recognition to identify a variety of things, including currencies, text, documents, and people, among others. One remaining challenge is automatic labeling and recognizing photos captured by blind users~\cite{gurari2020captioning}, which can be crucial for crowdsourcing-based assistive systems (e.g., Vizwiz~\cite{bigham2010vizwiz}) when crowd workers are not immediately available. 

Toucha11y also makes use of computer vision for two purposes. Toucha11y uses the classic SIFT algorithm~\cite{sift} to detect the key points from the photos taken by the onboard camera. It then matches the photos to the stocked touchscreen interfaces using FLANN~\cite{flann}, and utilizes the positions of these photos to compute the bot's actual location on the screen's X-Y coordinates.

\section{Formative Study}
To gain insights about blind users' experience with touchscreen-based kiosks (if any) and to identify key challenges, we conducted semi-structured interviews with nine blind individuals. Each interview was approximately thirty minutes long and was audio-transcribed. The design of Toucha11y was guided by the findings from the interviews and previous literature.

\subsection{Findings}
\subsubsection{Common practice}
While all interviewees reported using touchscreen-based personal devices (e.g., iPhones and Android smartphones) on a daily basis, none used public touchscreen-based kiosks regularly. Six of the interviewees noted that they had never used a public kiosk. For the remaining three, they reported a one-time experience of trying a kiosk on different occasions (to withdraw cash from an ATM, to make an appointment for a lab test in a clinic, and to renew an ID at the DMV), but because the experience was so negative, they no longer use kiosks by themselves. For all the participants we interviewed, when they needed to use a touchscreen-based kiosk, they either sought assistance from sighted people or chose not to use the device at all. 

\subsubsection{Barriers to using kiosks}
Two main challenges were identified as preventing blind interviewees from trying to use a public kiosk. 
First, participants noted that figuring out whether a public kiosk was accessible or not was challenging. Interviewees claimed that they were aware that kiosks were growing in popularity and that they had encountered them throughout their lives. Four participants also stated that they were aware that ADA regulations require that kiosks be accessible. However, with a huge percentage of existing kiosks being inaccessible, \textit{"the common sense for blind people when we come across a device with a touchscreen is that it doesn't work for us."} Thus, blind users who wish to use one must make additional efforts to first figure out whether or not it is usable.

Second, even kiosks that claimed to be accessible could be difficult to learn and use. One participant described their arduous experience withdrawing cash from an ATM kiosk. \textit{"I was told that the ATM was accessible, so I wanted to give it a chance... I found the ATM, looked for the headphone jack on the machine, and plugged my headphones into it. Although the bank claimed that I could navigate the screen to check my balance or withdraw money, there was no clear instruction on how to do it after I plugged in the headphones."} The interviewee ended up spending an hour figuring out how to use the touchscreen ATM on their own because no one was in the lobby to assist them. \textit{"It wasn't a very good experience, and that didn't work out. So at last, I gave up and left the bank."}   

Because blind participants had either experienced or were aware that many touchscreen-based kiosks were not accessible, they concluded that learning to use them would not pay off. As one participant put it, \textit{"if a sighted person could use a touchscreen in 30 seconds to finish the job, but I need to spend 10 minutes learning and using it, why should I?"} 

\subsubsection{Privacy concerns}
Since the use of kiosks was impractical for many blind users, the only viable alternative was to seek assistance from sighted individuals. However, when a blind individual is not accompanied by a friend or family member, seeking assistance from strangers could expose them to the risk of disclosing sensitive personal information. One participant stated, \textit{"There was a kiosk to let people register in the Social Security Office, but the machine wasn't accessible to me. I had to find someone on-site to help me, and I needed to tell them my social security number, which I didn't wish to do."} Other participants echoed this sentiment, \textit{"If I use a touchscreen-based ATM, will I even let others know my password? No, that isn't what I want."}

\subsection{Design Considerations}
The interview findings inform the three design considerations as stated below.
\begin{enumerate}
     \item \textit{Simplify the interaction}. As it remains difficult for blind users to figure out and learn how to use an arbitrary public kiosk due to the inconsistency of (existing) accessibility features, we will focus on simplifying the learning and interaction processes. Knowing that many blind users are active smartphone users, we aim to offload the interaction with a public kiosk to the user's personal smartphone device. The actual touch event on a touchscreen can be delegated to a robot that excels at repeatable and accurate touch actions.
    
    \item \textit{Protect privacy and promote autonomy}. Our solution shall protect the privacy of blind users and enable them to independently use touchscreen devices. For example, our solution shall allow blind people to enter their passwords or personal information without the assistance of sighted people.
    
    \item \textit{Save time}. Blind people may require more time than sighted people to interact with a touchscreen device they are unfamiliar with. Our solution shall be time-efficient. 

\end{enumerate}

\section{Toucha11y Walkthrough}

Toucha11y is a personal assistive device that enables blind people to use existing, inaccessible public touchscreen devices independently and with minimal effort. To better explain how Toucha11y works, in this section, we will walk through the use of Toucha11y in an example scenario where Alex---a blind user of Toucha11y---is attempting to order a bubble tea through an inaccessible public kiosk.

Alex is a blind user who enjoys bubble tea. Alex's favorite bubble tea shop recently began using a self-service kiosk for all food orders. Unfortunately, despite being brand new, the kiosk is touchscreen-only with no accessibility features, which Alex discovered only after arriving at the store. 

\begin{figure}[t]
  \includegraphics[width=\columnwidth]{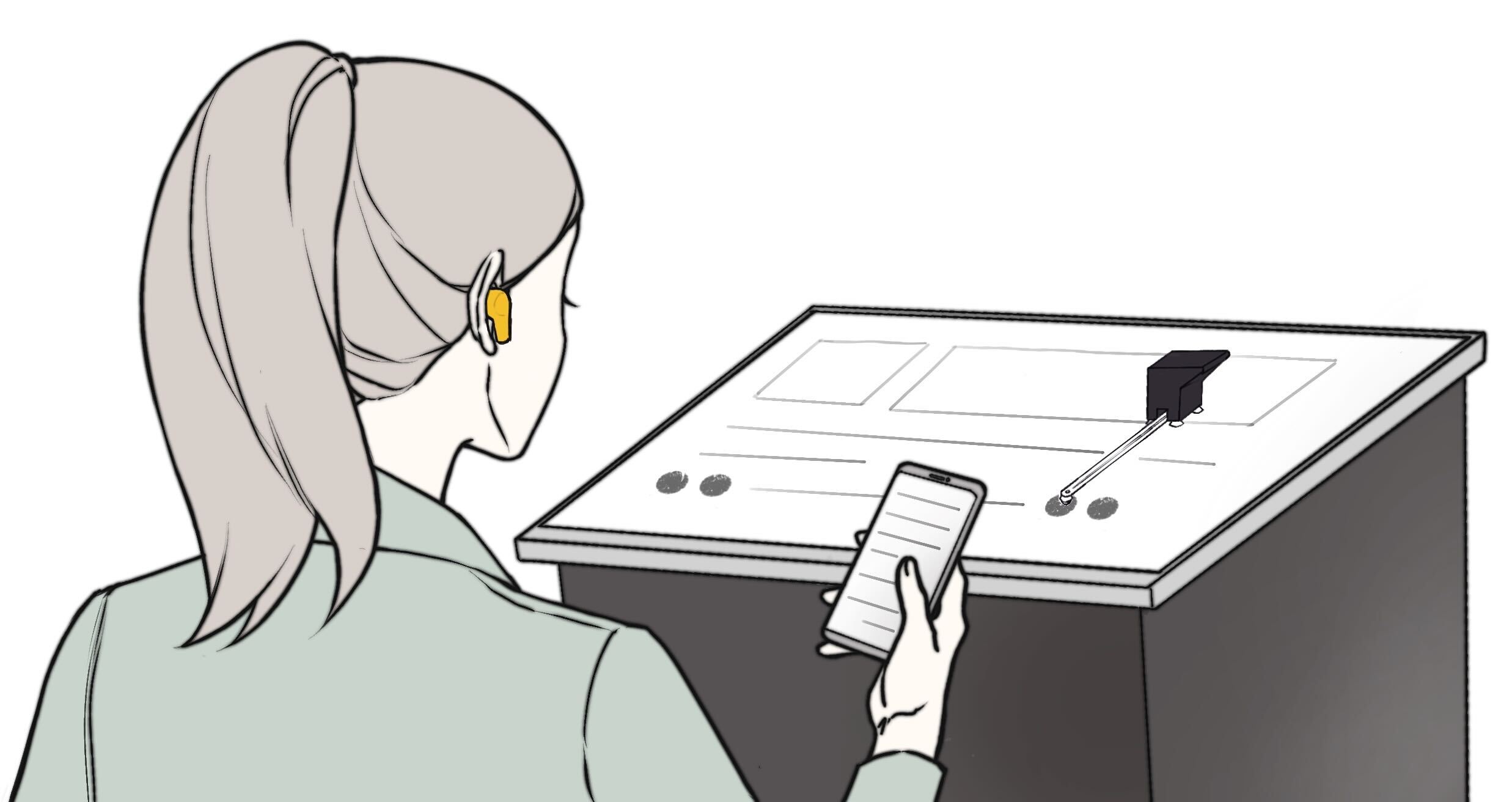}
  \caption{Using Toucha11y, Alex can access the touchscreen content from a smartphone.}
  \Description{Using Toucha11y, Alex can access the touchscreen content from a smartphone.}
  \label{fig:tea store}
\end{figure}

Although not ideal, with Toucha11y, Alex is able to order a drink at the self-serving kiosk (Figure~\ref{fig:tea store}). Alex first approaches the kiosk, locates its screen, and then attaches the Toucha11y bot to it. Confirming that the bot is securely in place, Alex launches the Toucha11y app on the smartphone. The bot begins to rotate and photograph the screen, allowing it to recognize the touchscreen context as well as its own location in relation to the touchscreen; at the same time, the mobile app retrieves and displays the corresponding menu of the bubble tea shop on Alex's smartphone. Instead of physically touching the inaccessible kiosk screen, Alex can use the phone to browse the menu using the smartphone's built-in screen reader features. 

Alex selects "Avocado Tea" and presses the "Add" button on the smartphone. The bot then activates the avocado tea button on the physical touchscreen, bringing the ordering process to the next step, tea customization. At the same time, Toucha11y the app prompts Alex with the same customization options. Alex makes their choice and presses the "Add to Cart" button on the phone. Meanwhile, the Toucha11y bot performs these actions on the physical touchscreen, completing Alex's order.

\section{Toucha11y}

\begin{figure*}[ht!]
  \includegraphics[width=\textwidth]{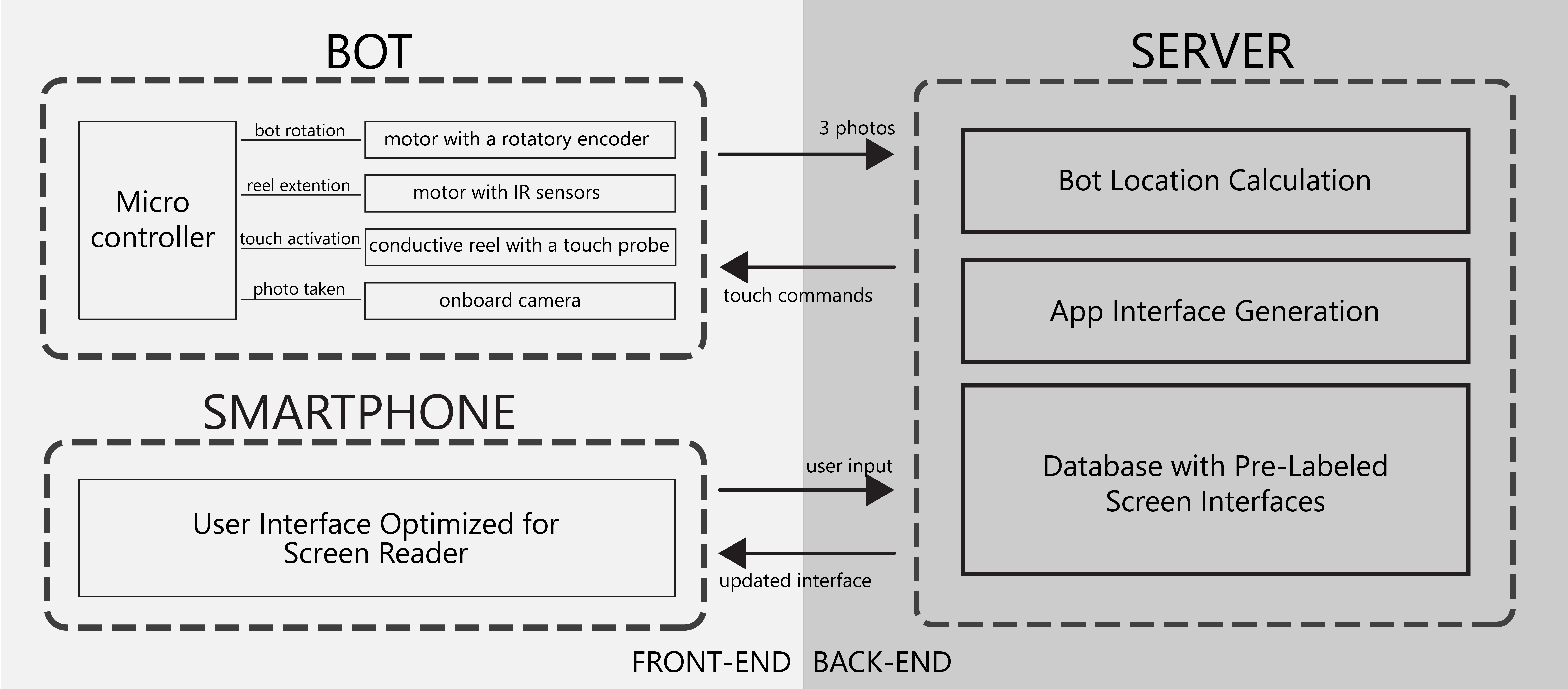}
  \caption{Toucha11y system architecture.}
  \Description{Toucha11y system architecture. The system has three parts, the bot, the smartphone, and the server. The bot and smartphone are the front-end, and the server is the back-end. The bot contains a microcontroller, a motor with a rotatory encoder to control bot rotation, a motor  with an IR sensor to control reel extension, a conductive reel with a touch probe to active touchscreens, and an onboard camera to capture photos. The smartphone has a user interface optimized for screen readers. The server has three main tasks, including bot location calculation, app interface generation, and a database with pre-labeled screen interfaces.}
  \label{fig:workflow}
\end{figure*}

Figure~\ref{fig:workflow} shows the Toucha11y architecture, which consists of a mechanical bot that physically registers touch events for a blind user, a smartphone app that the user can interact with, and a back-end server that bridges users' input and the bot's actuation.
Note that since previous work, such as Statelens~\cite{guo2019statelens}, has already detailed how to reverse engineer and label an arbitrary touchscreen interface, we assume that the touchscreen interface information can be directly retrieved from the server. 
Built upon, Toucha11y's main focus is a tangible solution that allows blind users to access public touchscreens with a simple interaction model, less concern for privacy, and independence. 

Below we detail each of the Toucha11y system components.

\subsection{The Toucha11y Bot}
\begin{figure}[b!]%
  \includegraphics[width=0.99\columnwidth]{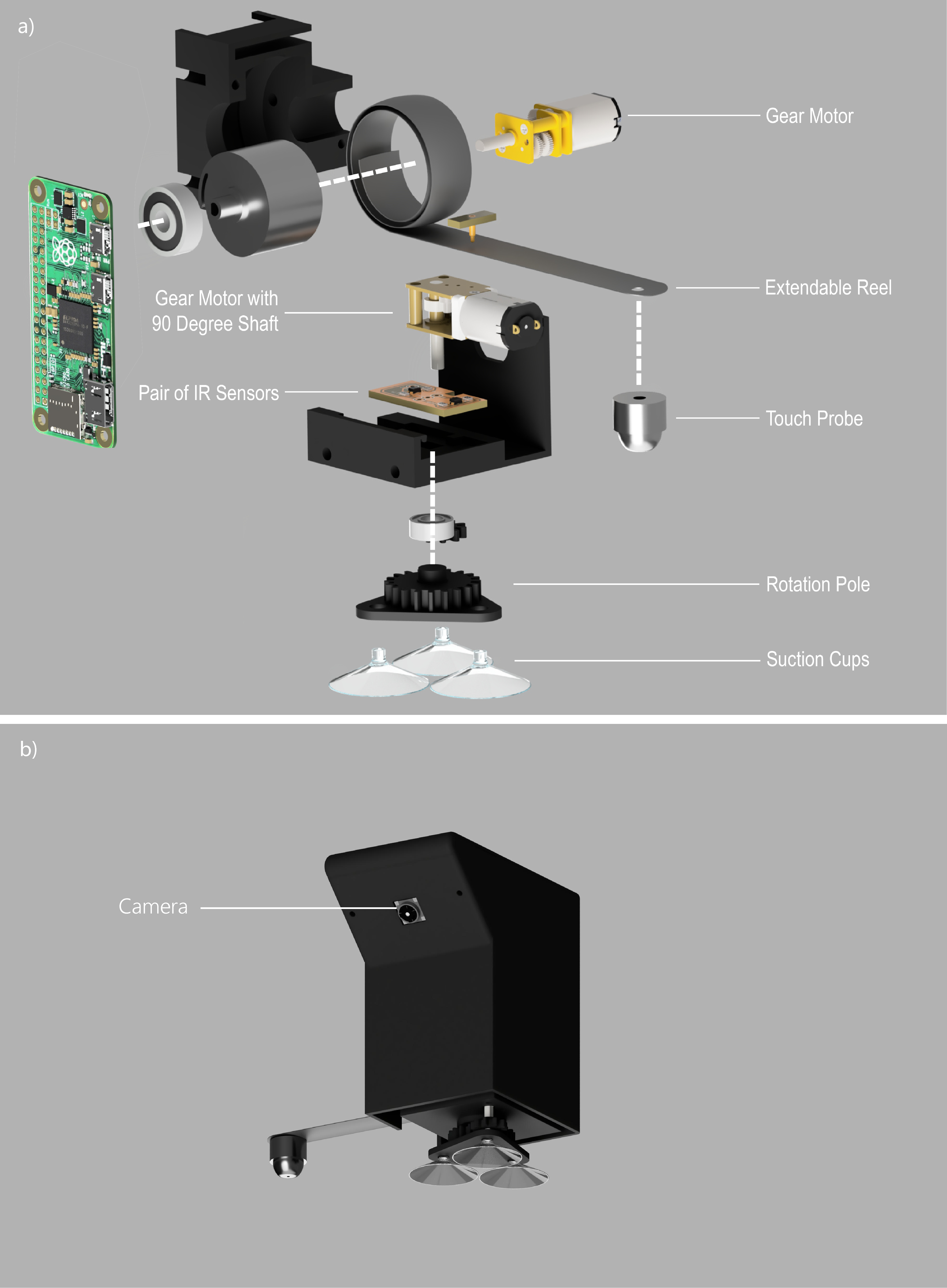}
  \caption{Toucha11y assembly illustration.}
  \Description{Toucha11y assembly illustration.}
  \label{fig:assembly}
\end{figure}

Figure~\ref{fig:assembly}a is an exploded view of the bot design---the actuator of the Toucha11y system that physically registers touch events for a blind user. 
To accomplish this, the bot must be able to: 1) fix to a touchscreen surface; 2) locate itself on the touchscreen; and 3) register touch events according to the user's input. 

\subsubsection{Fix to a touchscreen surface}
The first step to using Toucha11y is to fix the bot to a touchscreen kiosk. To ease the anchoring process for blind users, the base of the Toucha11y bot is equipped with three \SI{19}{\milli\metre} diameter suction cups spaced \ang{120} apart (Figure ~\ref{fig:reel}b). Three suction cups will guarantee the bot's overall stability while it physically activates the screen; the redundancy in the number of suction cups is to ensure that the device can still stay on the screen even if one of the suction cups fails.

\subsubsection{Photograph the touchscreen for localization}\label{localization}
Once the bot is fixed on the touchscreen, it needs to recognize its precise placement with respect to the screen's coordinates in order to register touch events. 
Toucha11y bot accomplishes this by taking three consecutive photos of the screen interface beneath (each with a \ang{30} gap), which, after each shot, are uploaded to the back-end server. Since the bot's microcontroller, the Raspberry Pi Zero W, is not suitable for heavy computation, the on-screen localization is done on the server using off-the-shelf computer vision algorithms SIFT~\cite{sift} and FLANN~\cite{flann}, which we will briefly explain in Section~\ref{algorithm}. 

To take photos of the screen, the bot is equipped with a Pi V2 camera (8 megapixels, 1080p), mounted \SI{70}{\milli\metre} above the bot's base and angled \ang{45} downward (Figure~\ref{fig:assembly}b). 
This configuration is to ensure that each photo captured by the camera can cover a large screen area, in this case, \SI{120}{\milli\metre\squared}. Currently, it takes the bot four seconds to capture all three photos of the screen.

\subsubsection{Touch event registration}
After locating itself on a touchscreen, the bot is ready to register touch events mechanically. The bot design employs two-dimensional polar coordinates where the main body of the bot rotates along the origin and an extendable reel can change length along the polar axis (Figure~\ref{fig:assembly}b). Such design makes it possible for the extendable reel to reach virtual screen buttons that are far from the bot's body while remaining completely hidden when not in use, thereby reducing the overall size of the bot.

\begin{figure}[t]
  \includegraphics[width=\columnwidth]{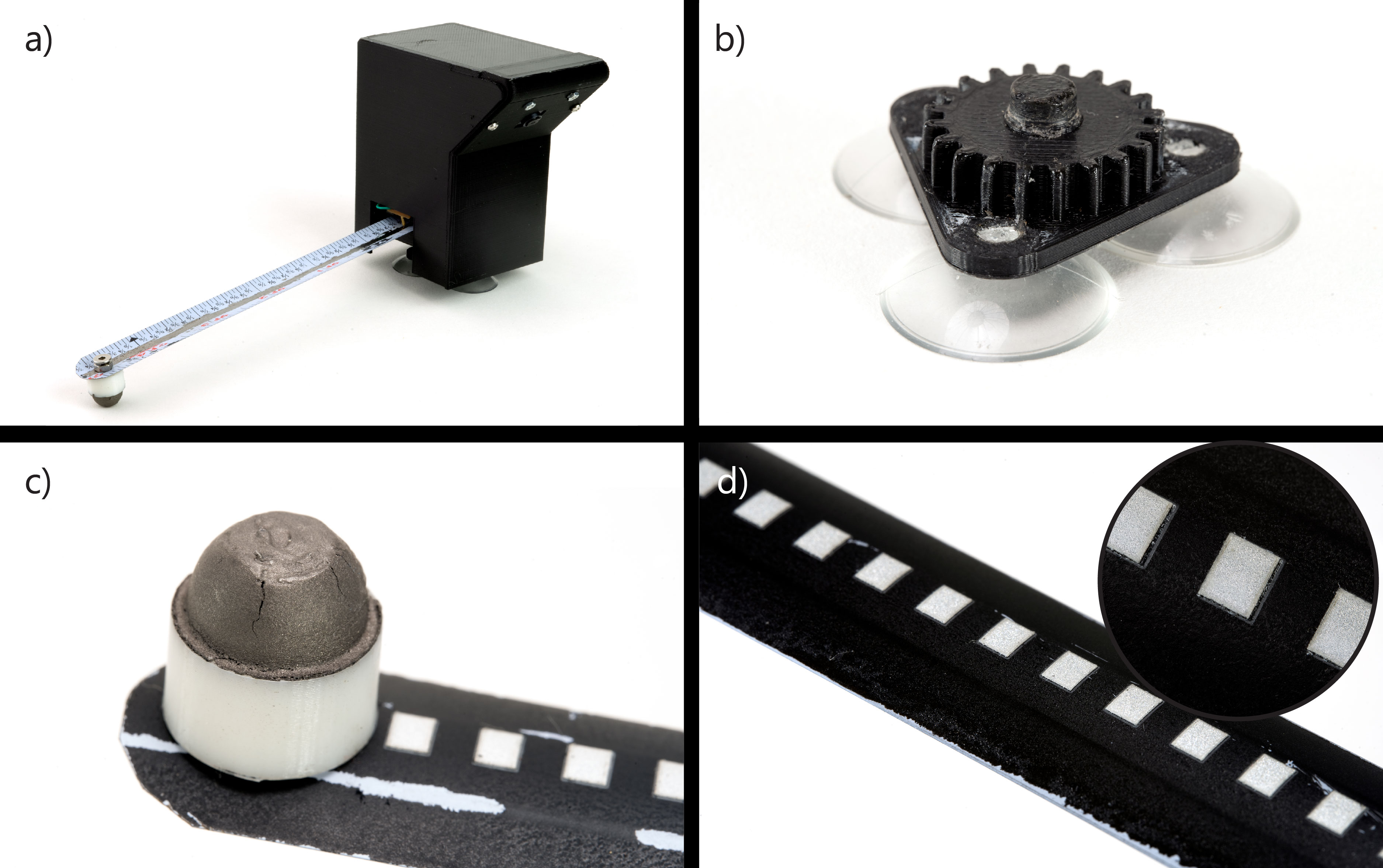}
  \caption{a) The Toucha11y bot. b) The bot base with three suction cups. c) A 3D-printed touch probe coated with conductive material. d) Extendable reel made from stainless steel measuring tape with black backing and reflective white stripes. }
  \Description{a) The Toucha11y bot. b) The bot uses three suction cups to attach to screens. c) A 3D-printed touch probe coated with conductive material. d) Extendable reel made from stainless steel measuring tape with black backing and reflective white stripes. }
  \label{fig:reel}
\end{figure}

The extendable reel design is inspired by ~\cite{suzuki2019shapebots}. The reel is made from a portion of a stainless steel measuring tape (Figure~\ref{fig:reel}a). 
Through experiments, we found that the stainless steel reel is both rigid and lightweight, allowing it to remain straight even when extended over a great distance (up to \SI{700}{\milli\metre} with our current prototype).
The far end of the extendable reel is installed with a 3D-printed touch probe (coated with conductive paint, MG Chemicals MG841AR) facing the touchscreen surface (Figure~\ref{fig:reel}c). When a virtual button on a touchscreen needs to be activated, the reel first extends out from the bot body at an angle of about \ang{5} pointing downward (Figure~\ref{fig:reel}a). Once the touch probe reaches the target, a touch event signal from the bot's microcontroller is transmitted through the stainless steel reel to the touch probe, which then activates the touchscreen interface. 
Note that the slight tilting angle of the extendable reel is to ensure that the touch probe always has secure contact against the touchscreen across the entire area. Also, as the touch mechanism is activated electrically, a touch event will only be triggered once the probe arrives at the prospective location.

To ensure that the length of the extrusion is accurate, the back side of the reel is painted black with reflective white stripes evenly spaced (at a pitch of \SI{2.5}{\milli\metre}), which are detected by a pair of IR sensors for length counting (Figure~\ref{fig:reel}d).
The maximum reach of the current prototype is \SI{700}{\milli\metre}, which is enough to encircle a 40-inch touchscreen device if placed at its center. The extrusion speed of the reel is \SI{25}{\milli\metre/\second}; the rotation speed is \SI{5}{rpm}.

\subsubsection{Implementation}
The bot prototype is self-contained, with an overall size of \SI{50}{\milli\metre} by \SI{70}{\milli\metre} by \SI{93}{\milli\metre} and a weight of \SI{162.53}{\gram}. The bot is equipped with a Raspberry Pi Zero W as the microcontroller, two N20 gearmotors to drive the rotation and the reel extension, and a \SI{7.4}{\volt}, \SI{260}{mah} LiPo battery as the power source.
Given that the controller only accepts \SI{5}{\volt} as input voltage, we use an MP1584 DC-DC step-down converter to regulate the voltage. 
Additionally, the battery powers the two motors through a dual H-bridge DRV8833. 
The pole rotation motor is equipped with a magnetic rotary encoder and a \ang{90} shaft that connects to the suction cup base. The motor of the extension reel is positioned orthogonally above (Figure ~\ref{fig:assembly}a). Two IR sensors (SHARP GP2S60) are positioned beneath the extendable reel to measure extension length.
The rotation angle of the bot and the length of the extendable reel are both computed using a custom PID implementation.

\begin{figure}[t]
  \includegraphics[width=\columnwidth]{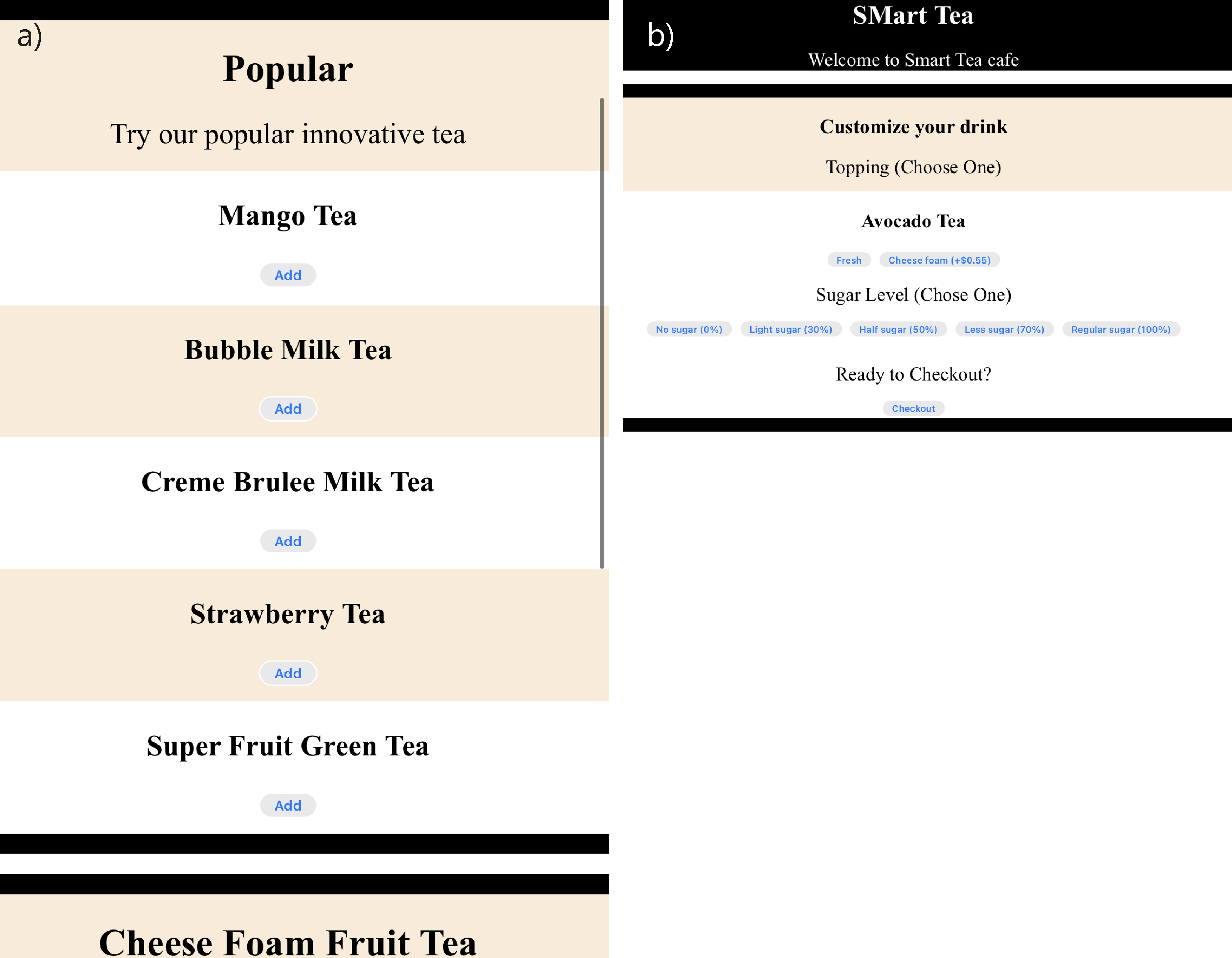}
  \caption{An example of the smartphone interface for blind users.}
  \Description{An example of the smartphone interface for blind users.}
  \label{fig:smartphone interface}
\end{figure}

\subsection{The Smartphone Interface}
The main goal of the smartphone interface is to allow blind users to access and interact with the content of a public kiosk using their own equipment, which not only has accessibility features built-in but can also mitigate privacy concerns. 

In the current implementation, the HTML-based smartphone interface is automatically generated on the server and pushed to the user's phone (see Section~\ref{interface generator} for interface generation). Figure~\ref{fig:smartphone interface} showcases one example of a bubble tea menu being converted to a smartphone interface. Note that all touchable elements are marked up in HTML according to WCAG standards~\cite{w3c} so that they can be verbalized by a screen reader.

\subsection{The Server} 
While the back-end server is not directly accessed by end users, it has three key functions that bridge the bot actuation and user input. In particular, the Toucha11y server stores the pre-labeled touchscreen interface information in a database, computes the placement of the bot, and generates the smartphone interface for the user. It also sends touch commands to the bot based on the user's selection from their smartphone, and updates the phone interface whenever the touchscreen is updated (Figure ~\ref{fig:workflow}). 

Flask~\cite{flask} is used to implement the server, which is deployed on the Heroku online cloud platform. The server communicates with both the bot and the user's smartphone wirelessly. 

\begin{figure}[t]
  \includegraphics[width=\columnwidth]{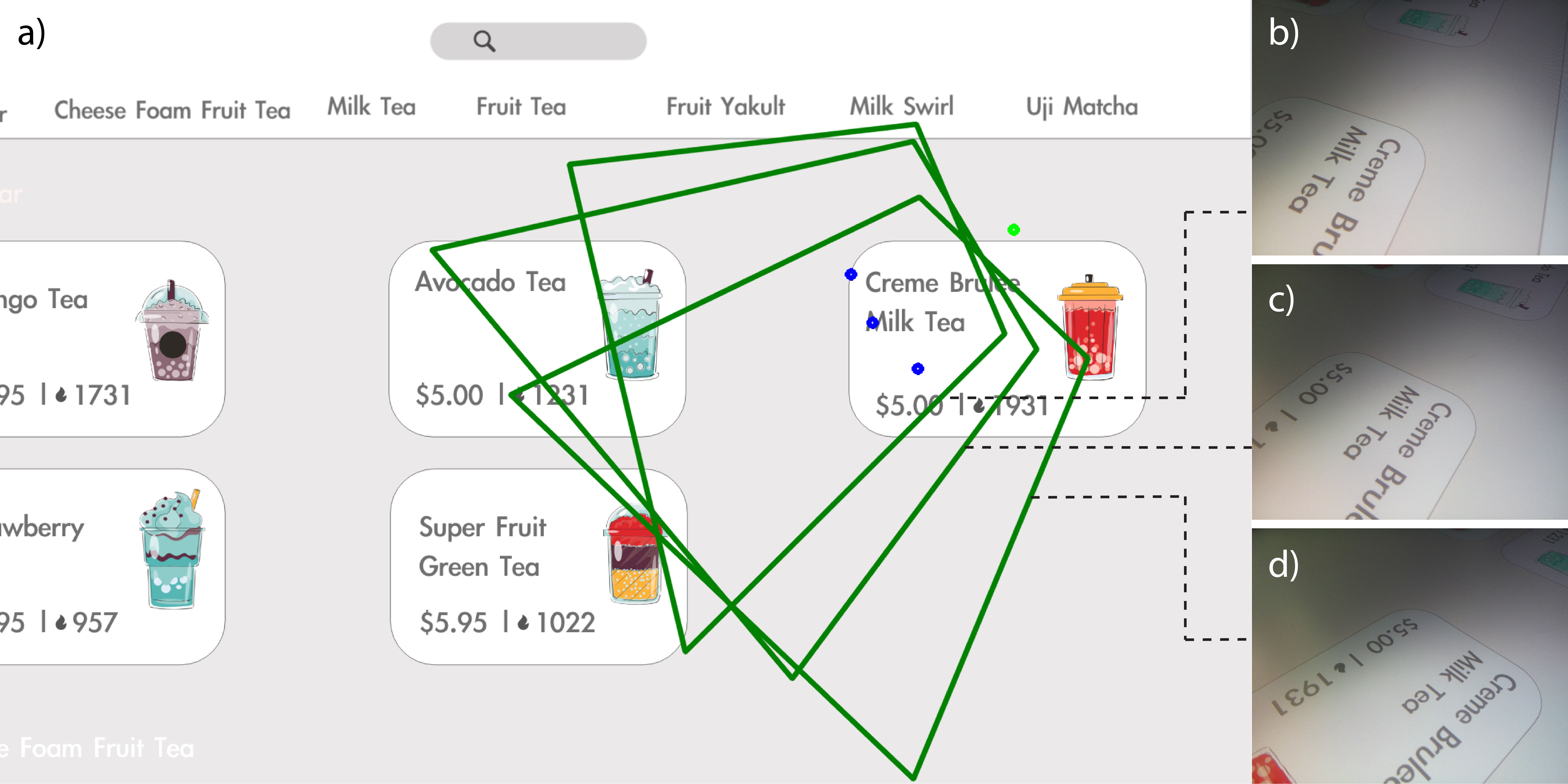}
  \caption{a) The green boxes represent the perspective-transformed area covered by the photos taken by the onboard camera, shown as b), c), and d). The green dot shows the estimated location computed from the three blue dots, which are the perspective-transformed points corresponding to the photos' center coordinates.}
  \Description{a) The green boxes represent the perspective-transformed area covered by the photos taken by the onboard camera, shown as b), c), and d). The green dot shows the estimated location computed from the three blue dots, which are the perspective-transformed points corresponding to the photos' center coordinates.}
  \label{fig:sift}
\end{figure}

\subsubsection{Database}\label{database}
A simple, customized database is hosted on the server and stores the essential information of the public touchscreen interfaces. Specifically, for each instance of a touchscreen interface, three types of data are stored, including all the text contents on the interface, their click-ability, and their location coordinates on the screens. The first two are used to generate the smartphone interface, and the location coordinates are sent to the bot for touch activation. Additionally, one image per interface is stored in the database for computing the bot's location. Note that the interfaces in the database are manually labeled for the current prototype, but we assume that they can be labeled via crowdsourcing if deployed in the future, as discussed in ~\cite{guo2019statelens}.

\subsubsection{Location computation}\label{algorithm}

The server determines the location of the bot based on the three photos taken by the bot's onboard camera and the stored kiosk interface images in its database. The computation can be broken down into three steps: finding the matching interface image, computing the locations of the camera photos in the interface image coordinates, and then triangulating the bot's location. 

To determine which touchscreen kiosk the bot is placed on and its corresponding interface image in the database, the server first uses SIFT~\cite{sift} to generate a list of key points (represented as vectors) in the camera photos. 
These key points are then compared to those of the pre-labeled interface images in the database. This is achieved using FLANN~\cite{flann} by finding the nearest neighbors of the interface image's key points that match those of the camera photos; the one with the highest number of matching key points is regarded as the touchscreen interface on which the bot is placed on. Once the interface image has been identified, the server uses a perspective transformation to compute the locations of all three camera photos within its coordinates. For example, the three green boxes in Figure~\ref{fig:sift} represent the transformed area of the three camera photos, whereas the three blue dots mark the geographical center of the camera photos after the transformation.
A circle that traverses the transformed center coordinates is then calculated. 
As all three photos are taken from the same bot's camera rotating along its pole, the circle center (the green dot in Figure~\ref{fig:sift}a) marks the exact location of the bot.

\begin{figure}[t]
  \includegraphics[width=\columnwidth]{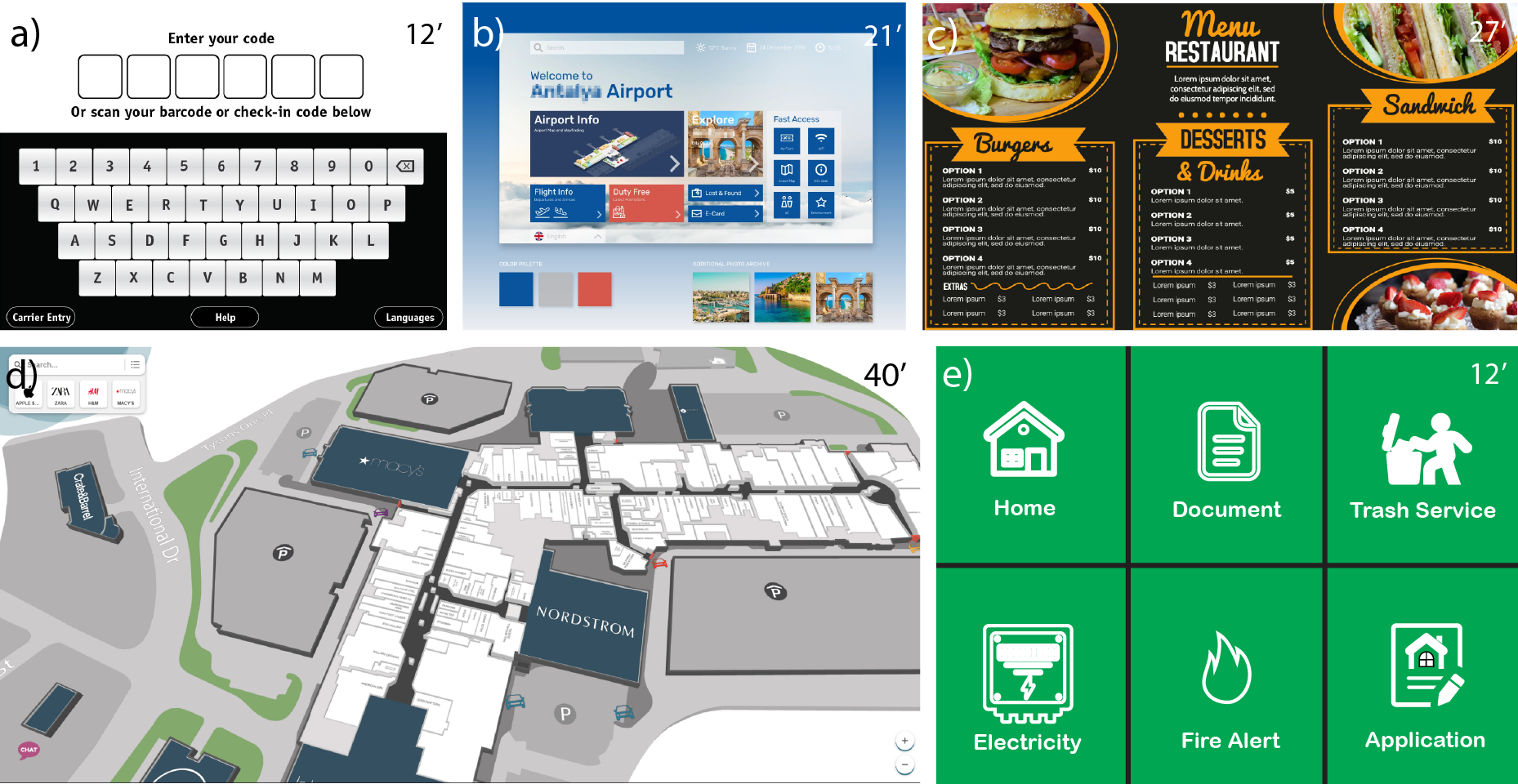}
  \caption{Samples of touchscreens' interfaces.}
  \Description{Samples of touchscreens' interfaces. a) Self-serving mailbox kiosk user interface, b) airport touchscreen user interface, c) restaurant serf-ordering kiosk user interface, d) map kiosk, e) single color background touchscreen user interface.}
  \label{fig:interface samples}
\end{figure}

\subsubsection{App interface generation}\label{interface generator}
The smartphone interface is automatically generated on the server and updated on the user's smartphone. In particular, the interface is generated as an HTML page based on the touchscreen interface being identified. The HTML page renders a list of textboxes and buttons based on the pre-labeled interface information in the server's database and is sent via WebSocket. Once the bot activates a virtual button on the touchscreen of a kiosk and its interface changes, a new HTML page will be generated on the server accordingly and sent to the user's phone.

\section{Technical Evaluation}
Toucha11y is essentially an open-loop system, meaning that whether it can help blind users successfully activate a touchscreen device depends on its own accuracy of localization, pole rotation, and reel extrusion. 
Thus, to understand the performance of our current working prototype, we conducted three technical evaluations.

\subsection{Location Accuracy}
The goal of this evaluation is to determine whether the Toucha11y bot can locate itself on touchscreens of various sizes and configurations. We collected five public kiosk interfaces, as shown in Figure~\ref{fig:interface samples}. We began with four kiosk interface examples, each with a screen size of 12 inches, 21 inches, 27 inches, and 40 inches, representing a variety of use cases, including public lockers, airport kiosks, restaurant menus, and shopping mall navigation kiosks. As our algorithm is based on the number of visual features that can be detected on an image, we included a fifth interface example, a 12-inch touchscreen interface with a simple visual design and a monochrome background, to understand the prototype limitation.

\subsubsection{Procedure}
We located five testing points for each of the five touchscreen interfaces, four of which were located near the screen's four corners and one at the center. We assumed that blind users would find it easier to locate a kiosk device's physical bevel; thus, the corners might serve as physical references for blind users when placing Toucha11y devices. Similarly, the center of a screen might provide a sufficient flat surface area for blind users to work with.

\begin{figure}[b]
  \includegraphics[width=\columnwidth]{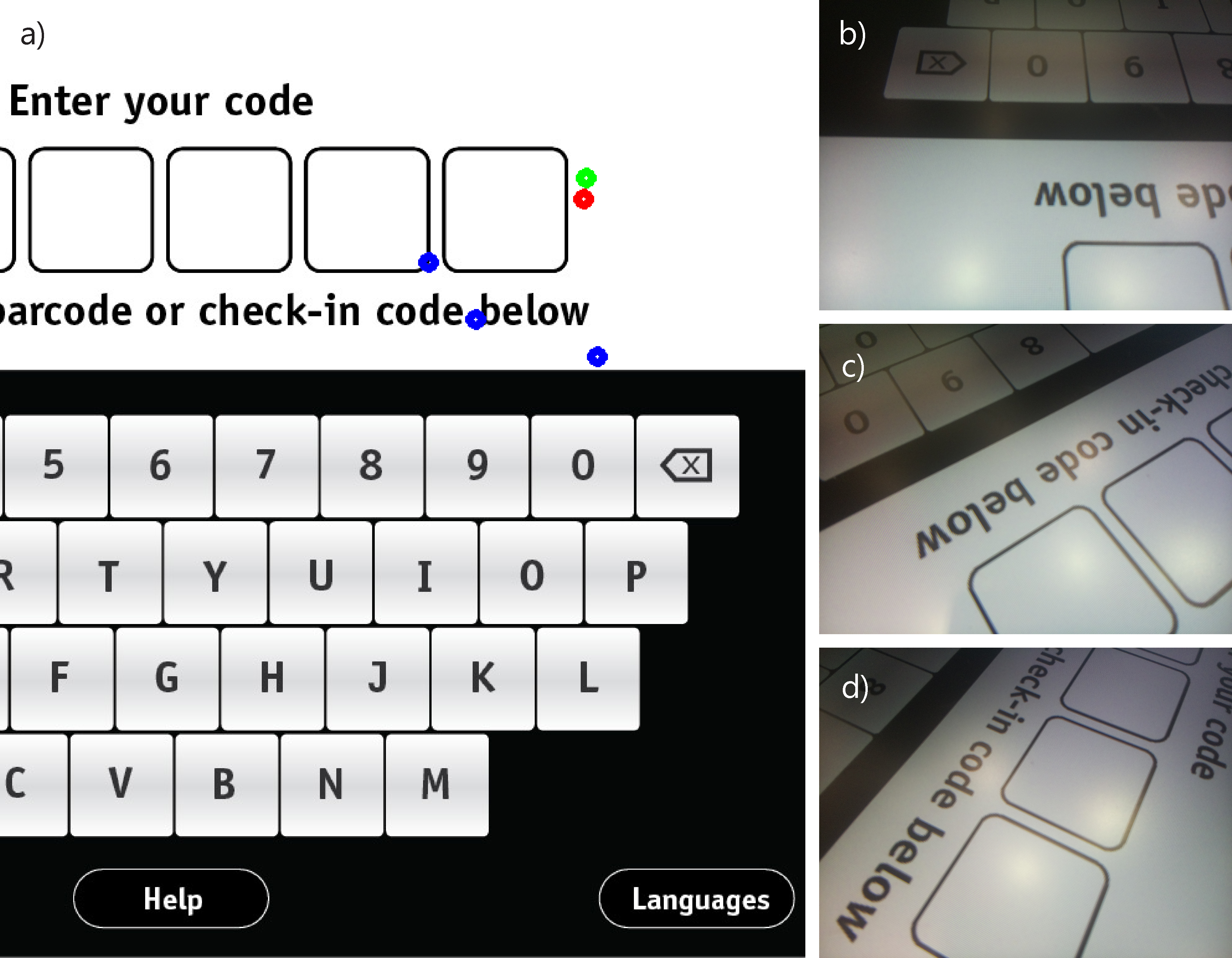}
  \caption{a) The red dot represents the actual location of the bot on the display, whereas the green dot represents the estimated location computed from the three blue dots based on the photos taken by the onboard camera, shown as b), c), and d).}
  \Description{a) The red dot represents the actual location of the bot on the display, whereas the green dot represents the estimated location computed from the three blue dots based on the three images shown as b), c), and d).}
  \label{fig:calculate location}
\end{figure}

For each testing point, we first created a red circular mark at the desired screen location. We then placed the Toucha11y bot right above the red mark. The bot would then proceed with the localization procedure as outlined in section~\ref{localization}, i.e., taking three consecutive photos of the screen with a \ang{30} interval in between. The three photos were used to estimate the bot's location, as shown in Figure~\ref{fig:calculate location}. Each testing location was evaluated three times, resulting in 15 data points for each screen interface example. We computed the error in the distance between the estimated and the actual coordinates for each data point.

\begin{table}[]
\caption{The distance between the calculated location and the actual location (unit: mm)}
\label{tab:localization}
\begin{tabular}{@{}llllll@{}}
\toprule
        & Sample 1 & Sample 2 & Sample 3 & Sample 4 & Sample 5 \\ \midrule
Point 1 & 3.56      & 4.88     & 7.54     & 6.21     & N/A     \\
Point 2 & 6.08      & N/A     & 4.06      & 8.72     & N/A     \\
Point 3 & 4.66      & 10.39      & 5.21    & 7.01     & N/A     \\
Point 4 & 4.11      & 11.98     & 6.12    & 10.02     & N/A    \\
Point 5 & 6.15      & 3.92     & 3.23     & 8.32     & N/A    \\
Mean    & 4.91      & 7.79     & 5.23     & 8.06     &  N/A    \\ \bottomrule
\end{tabular}
\end{table}

\subsubsection{Result}
\begin{figure}[t]
  \includegraphics[width=\columnwidth]{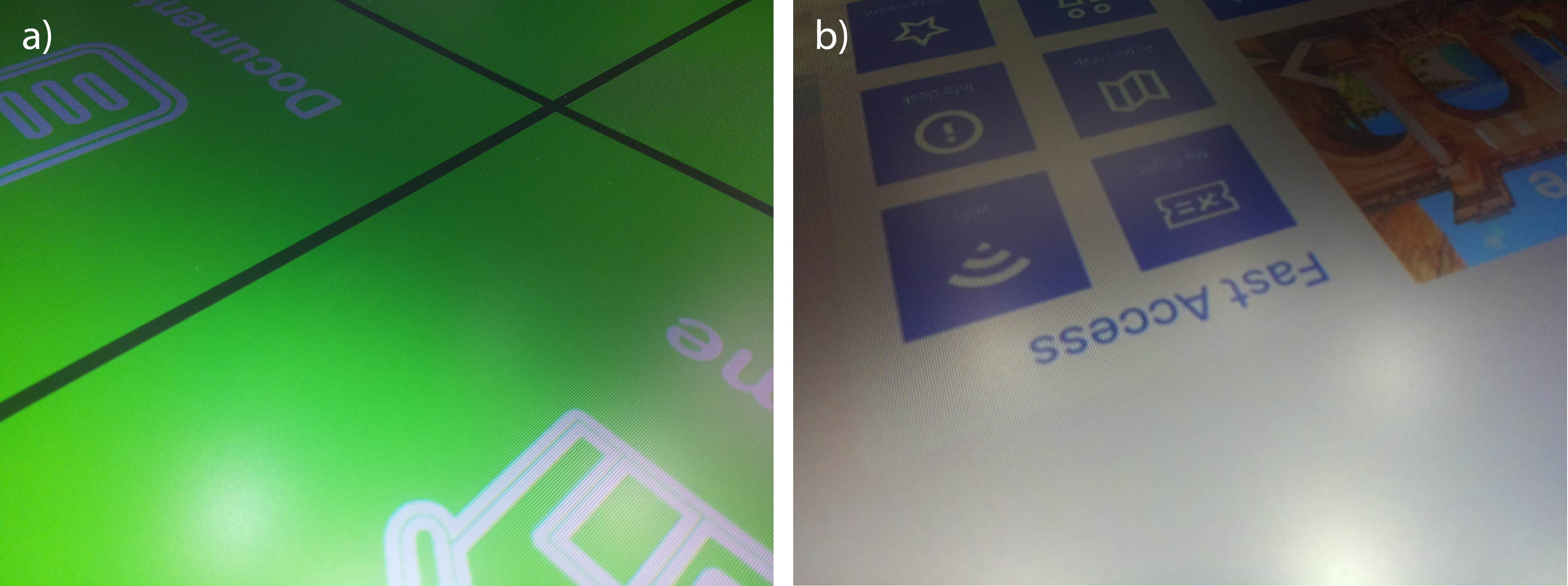}
  \caption{Toucha11y could not localize itself when the captured image has insufficient visual features.}
  \Description{Toucha11y could not localize itself when the captured image has insufficient visual features.}
  \label{fig:Sample 2 point 2}
\end{figure}

Table~\ref{tab:localization} summarizes the distance error between the calculated location and the actual location. Overall, the Toucha11y bot can be used to locate itself on touchscreen interfaces with different sizes and contents by taking a minimum of three photos with its camera. With our sample interfaces, the Toucha11y calculated its location with an average error distance of \SI{6.50}{\milli\metre} (\textit{SD} = 1.66) for all but one point of the first four interfaces. The error distance is sufficient for many touchscreen interfaces, given that most touchscreen interface UIs will have buttons of the size of \SI{25}{\milli\metre} by \SI{25}{\milli\metre}~\cite{thumb_size}. For the failed case, i.e., the upper right corner of sample 2, as shown in Figure~\ref{fig:Sample 2 point 2}b, the bot was unable to compute its location due to the inefficient interface features being captured. This is identical to the situation with sample interface 5. When the touchscreen interface has fewer features across the screen, our current implementation is incapable of calculating its location. 

Because the localization is primarily based on the implemented algorithm, we anticipate that the accuracy can be improved with software updates. For example, the current location of the bot is based on the center of a circle calculated from three photos taken with the bot, which is prone to error if one of the photos has an incorrect matching location. By using additional camera photos for the circle center calculation, we can potentially reduce this error and increase the precision of localization. Additionally, we can experiment with different image matching algorithms (e.g., ~\cite{leutenegger2011brisk, bay2006surf}), which may improve the system robustness under various touchscreen interface renderings (e.g. Figure~\ref{fig:Sample 2 point 2}).

\begin{figure}[b]
  \includegraphics[width=\columnwidth]{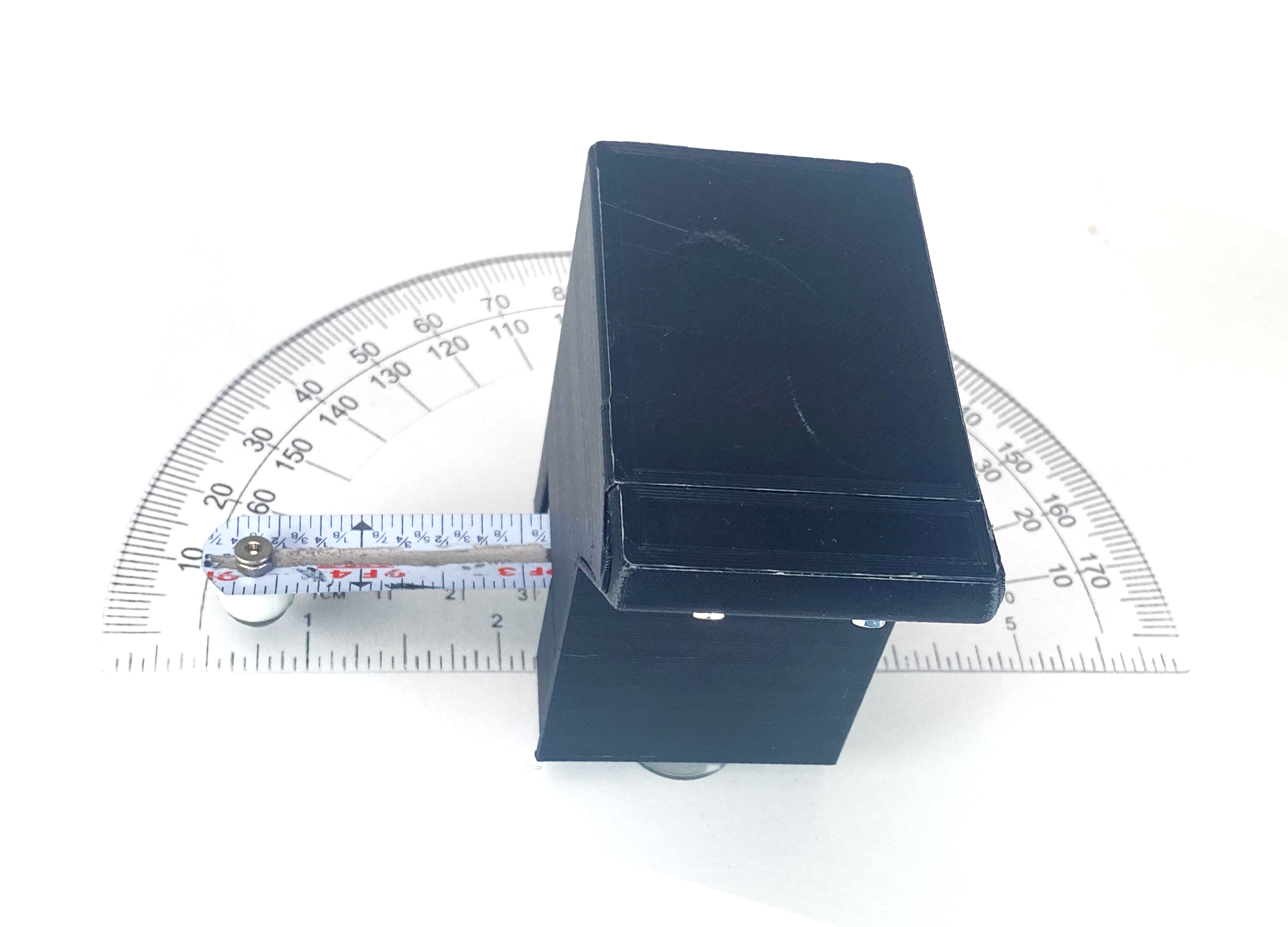}
  \caption{Rotational accuracy test with a protractor.}
  \Description{Rotational accuracy test with a protractor.}
  \label{fig:rotational_test}
\end{figure}
\subsection{Rotational Accuracy}
To determine the rotational accuracy of our prototype, we commanded the Toucha11y bot to rotate to various degrees and then measured the rotational error (Figure~\ref{fig:rotational_test}). In particular, the bot was first positioned in the center of a protractor aimed at angle \ang{0}. It then rotated to \ang{180} in \ang{30} increments, and back. We repeated the experiment three times for each angle and reported the average of the three experimental results, keeping two decimal places.

\subsubsection{Result}
Table~\ref{tab:rotation} reports the angular differences between the actual angle and the target angle in degrees. The average rotation error is 0.66 degrees (\textit{SD} = 0.60). The error is primarily caused by the small backlash from 3D-printed gears and human manufacturing errors. The error can be further reduced with enhanced manufacturing and assembly processes.

\begin{table}[]
\caption{The angular difference between the actual angle and the target angle (unit: degree)}
\label{tab:rotation}
\begin{tabular}{@{}llllllll@{}}
\toprule
                                                             & 0       & 30    & 60    & 90    & 120   & 150   & 180   \\ \midrule
Clockwise                                                    & 0       & 0.17 & 0.53 & 0.07 & -0.30 & -0.30 & -0.60 \\
\begin{tabular}[c]{@{}l@{}}Counter \\ Clockwise\end{tabular} & -0.67 & 0.23 & -0.57 & -0.07 & 0.10 & 0.40 & 0   \\ \bottomrule
\end{tabular}
\end{table}

\begin{table*}[]
\caption{Participants demographics.}
\label{tab:my-table}
\resizebox{\textwidth}{!}{%
\begin{tabular}{@{}ccllllll@{}}
\toprule
\textbf{ID} &
  \textbf{Gender} &
  \multicolumn{1}{c}{\textbf{Vision Level}} &
  \multicolumn{1}{c}{\textbf{Hearing Level}} &
  \multicolumn{1}{c}{\textbf{\begin{tabular}[c]{@{}c@{}}Touchscreen Accessible \\ Aids Software \end{tabular}}} &
  \multicolumn{1}{c}{\textbf{\begin{tabular}[c]{@{}c@{}}Accessible Aids\\ Software Use\end{tabular}}} &
  \multicolumn{1}{c}{\textbf{\begin{tabular}[c]{@{}c@{}}Education \\ Background\end{tabular}}} \\ \midrule
  
P1  & Female & Blind & Good & Voice-over on iPhone & 11 Years & Bachelor \\
P2  & Female & Blind & Good & Voice-over on iPhone & 12 Years & Master \\
P3  &   Male & Low vision & Good & Voice-over on iPhone & 7 Years & Bachelor in progress \\
P4  & Female & Blind & Good & Voice-over on iPhone & 8 Years & Master \\
P5  & Female & Blind & Good & Voice-over on iPhone, Braille display & 4 Years & Master \\
P6  & Female & Low vision & Good & Voice-over on iPhone, Braille display & Not sure, but proficient & Master \\
P7  & Female & Blind & Good & Voice-over on iPhone & 9 Years & Two-year college \\ \bottomrule
\end{tabular}%
}
\end{table*}

\subsection{Extension Accuracy}
Finally, we evaluated the extension accuracy of the extendable reel. We instructed the bot to extend and retract in \SI{50}{\milli\metre} steps between 0 and \SI{700}{\milli\metre}. We ran three trials for each length and measured the actual length of extrusion. 

\subsubsection{Result}
The average error for the extension length is \SI{3.052}{\milli\metre} (\textit{SD} = 3.147) in both extension and retraction actions. 
The linear encoder pattern at the bottom of the reel is the main cause of the error. 
As we affixed the reel with reflective tape at \SI{2.5}{\milli\metre} spacing, the extension accuracy will not be greater than +-\SI{2.5}{\milli\metre}. Reducing the spacing will improve the extension accuracy.

\section{User Study}
We conducted a formative user study to evaluate how Toucha11y supports blind users in using an inaccessible touchscreen interface.

\subsection{Participants and Apparatus}
We recruited seven participants (six female and one male) through online postings. Five participants self-reported as blind; two were low vision. All participants were familiar with accessibility features such as VoiceOver or TalkBack.

\begin{figure}[ht!]
  \includegraphics[width=\columnwidth]{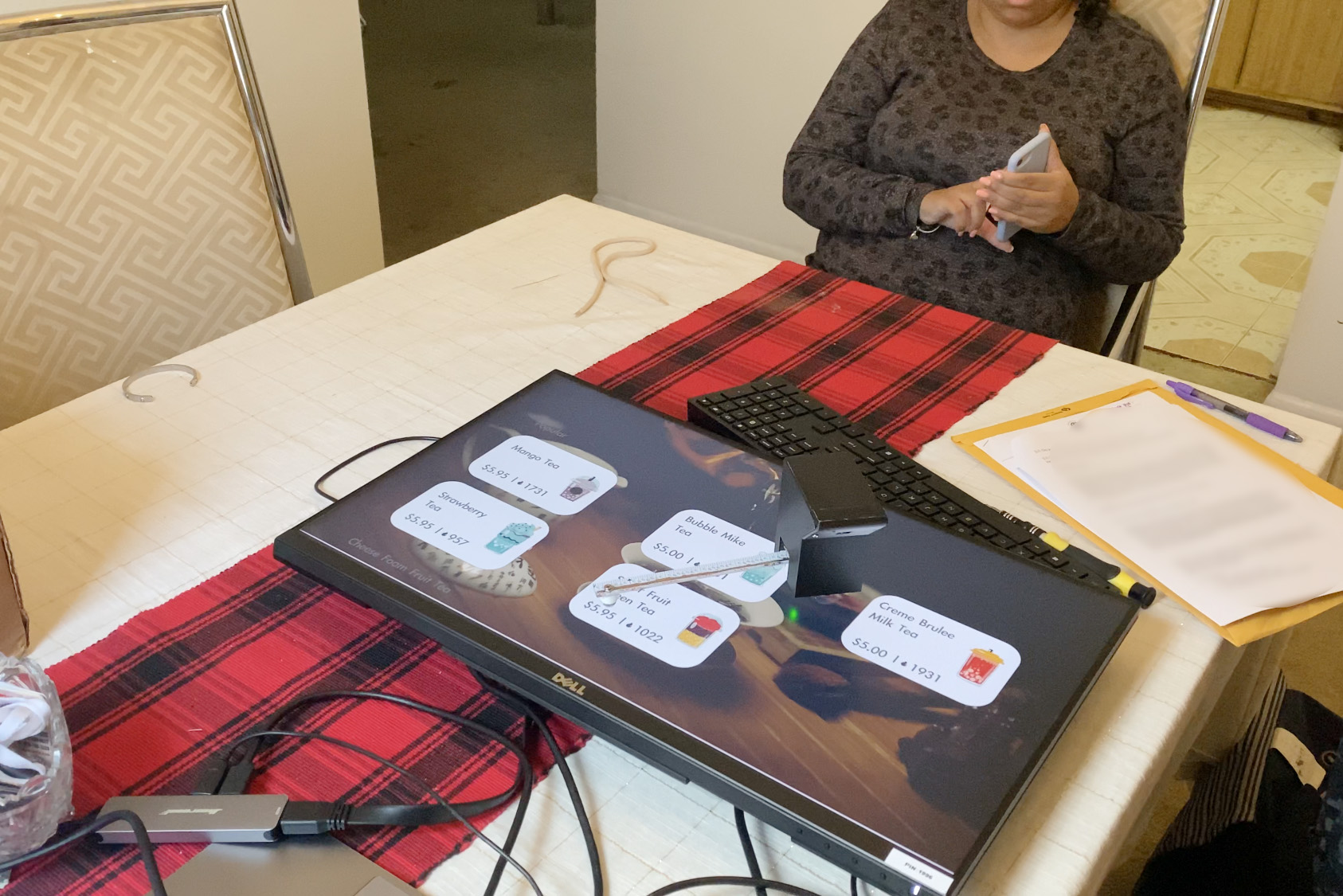}
  \caption{One participant is using a smartphone to explore the kiosk's touchscreen interface. The study was conducted at the participant's home.}
  \Description{One participant is using a smartphone to explore the kiosk's touchscreen interface. The study was conducted at the participant's home.}
  \label{fig:study procedure}
\end{figure}

The study apparatus included the Toucha11y bot prototype, an Android- or iOS-based smartphone depending on the preference of the participant, and a 24-inch touchscreen display (Dell P2418HT \cite{2418ht}). To simulate the use of a real touchscreen kiosk, we prototyped an interactive bubble tea menu using Adobe XD~\cite{xd}.

\subsection{Procedure}

\begin{figure}[b]
  \includegraphics[width=\columnwidth]{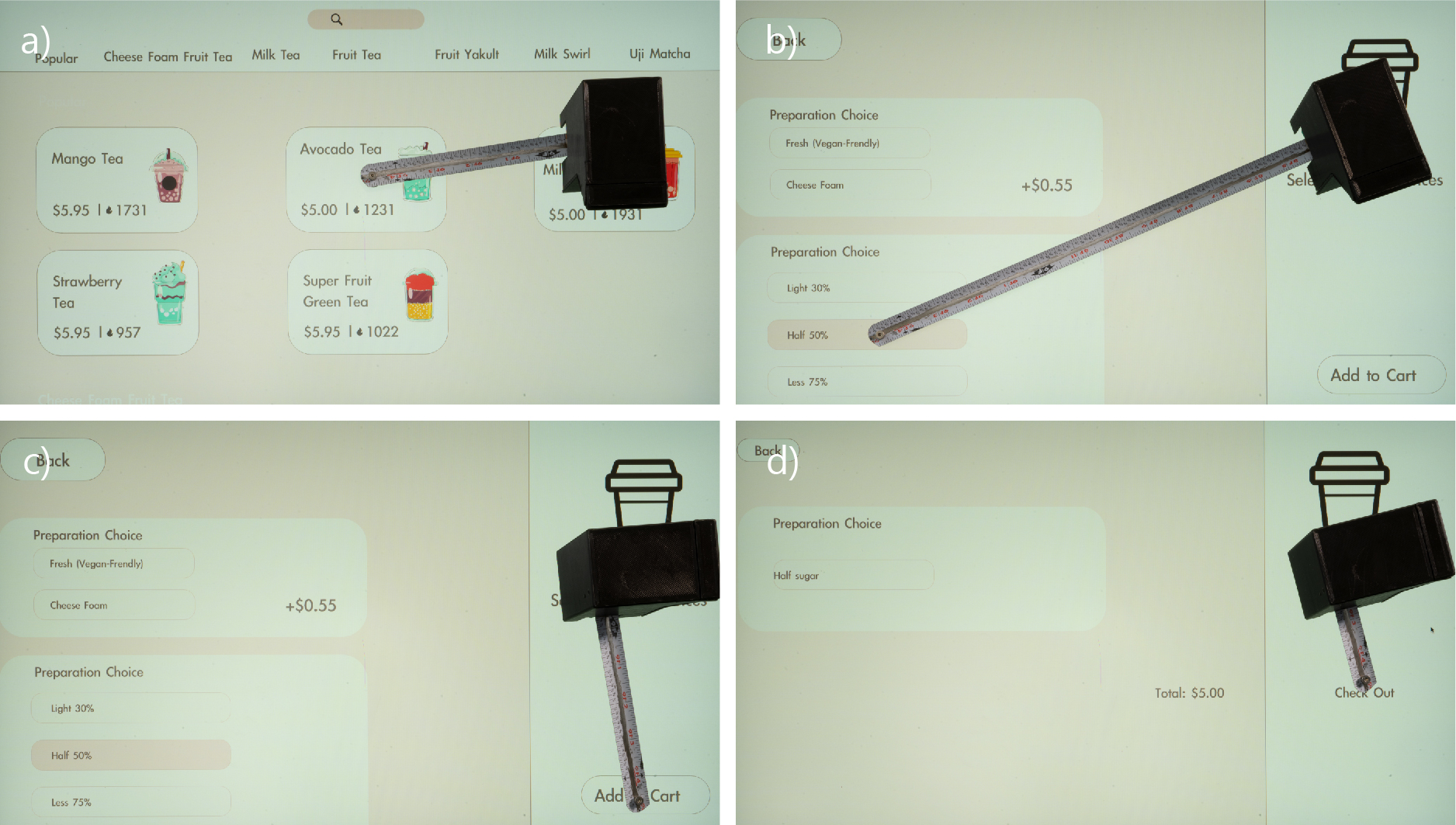}
  \caption{Study procedure. The blind users should make the purchase on their phone, with the Toucha11y bot completing the following actions: a) activate the "Avacado Tea" button, b) choose the half sugar option, c) triggers the "Add Cart" button, and d) "Check Out".}
  \Description{Study procedure. The blind users should make the purchase on their phone, with the Toucha11y bot completing the following actions: a) activate the "Avacado Tea" button, b) choose the half sugar option, c) triggers the "Add Cart" button, and d) "Check Out".}
  \label{fig:study procedure}
\end{figure}

\begin{figure*}[t]
  \includegraphics[width=\textwidth]{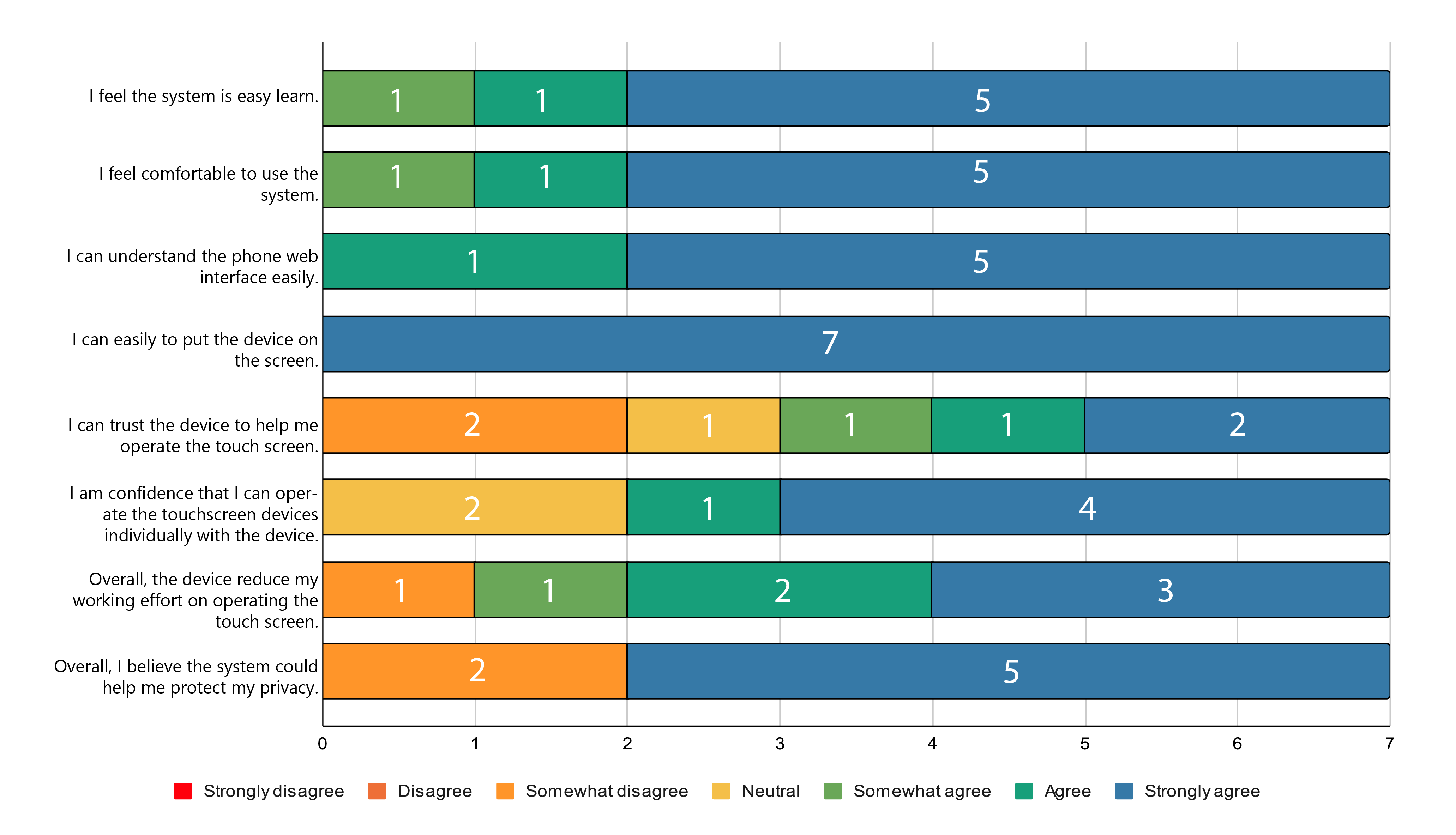}
  \caption{Self-reported ratings of using Toucha11y.}
  \Description{Self-reported ratings of using Toucha11y.}
  \label{fig:study result}
\end{figure*}

At the beginning of the user study, we collected the basic demographic information of the participants and then introduced the Toucha11y concept. 
Participants could ask any questions about the system and were encouraged to interact with the bot device, such as by touching it with their hands, putting it on the touchscreen display, or testing the smartphone app.
After participants felt comfortable and were familiar with Toucha11y, they were given the study task, which was to order an "avocado tea with 50\% sugar level" on an inaccessible touchscreen kiosk setup using the Toucha11y prototype.
Specifically, participants would need to 1) place the Toucha11y bot on the touchscreen, 2) find the correct tea options from their smartphone with in-system accessibility apps (e.g., VoiceOver), 3) choose the sugar level, and 4) confirm the order details and complete the transaction. Figure~\ref{fig:study procedure} shows the Toucha11y bot's activating the corresponding buttons on the touchscreen setup. 
To ensure consistency, all participants were asked to follow the same ordering routine. The duration of task completion was recorded.

Following the study, we first collected Likert scale ratings.
We then concluded with a semi-structured interview in which we solicited participants' comments and suggestions on the Toucha11y concept and the bot design. 
The research took about 40 minutes. 
Participants were paid at a rate of 30 USD per hour. 
For further analysis, the entire study was video- and audio-recorded.

\subsection{Results}

We present our user study result in this section and summarize the participants’ feedback. Note that the Likert scale questions are ranged from 1 to 7; 1 refers to strongly disagree, and 7 refers to strongly agree.

All participants were able to complete the tea ordering task successfully. The average time of completion was 87.2 seconds (\textit{SD} = 24.6). All of the participants found it easy to order with the smartphones, and most of the time was spent waiting for the bot to finish the button activation events. We observed that five of the participants simply placed the Toucha11y bot close to the touchscreen kiosk's edge or a corner. Two participants placed the Toucha11y bot in the central location. As their placements happened to block the touchscreen buttons that needed to be triggered, participants were prompted to relocate the bot to a new location to restart the task. The relocation time was not factored into the completion time calculation.

As shown in the summarized self-reported rating (Figure~\ref{fig:study result}), participants found that the Toucha11y device was easy to learn (\textit{M} = 6.57, \textit{SD} = 0.787), comfortable to use (\textit{M} = 6.57, \textit{SD} = 0.787), and very easy to put on the touchscreen (\textit{M} = 7.00, \textit{SD} = 0). The software interface provided sufficient instructions (\textit{M} = 6.71, \textit{SD} = 0.788). They were confident that with Toucha11y they could operate a public touchscreen kiosk independently (\textit{M} = 6.00, \textit{SD} = 1.41).

Participants believed that the device could reduce their cognitive effort (\textit{M} = 5.86, \textit{SD} = 1.46) and protect their privacy (\textit{M} = 5.86, \textit{SD} = 1.95) while enabling them to operate touchscreen-based devices independently (\textit{M} = 5.00, \textit{SD} = 1.73). The concerns of those participants who rated lower on these questions were mainly related to their lack of familiarity with the device and that the study was not in a real situation. As P6 said, \textit{"I am not sure I can trust this robot when I am in the grocery store alone, or if it can protect my privacy because you will not be there, and you can't teach me how to use this in the store... If I can practice more and be more familiar with the bot I could say yes...but right now I would say netural."}

Following the study, we solicited additional feedback on the Toucha11y concept from participants. All participants believed that the Toucha11y prototype could facilitate their use of touchscreen-based devices in public spaces. 
P7 stated, \textit{"I could bring that device to the Social Security Office and put it on the screen. It would take pictures of the information and transfer it to my phone."} P1 also envisioned a variety of situations that Toucha11y to be useful. \textit{"It can definitely help me at the doctor's office, airport, also DMV and other places. For my health insurance, I can use this on the company's kiosk, as it can help me check in and give me information, so I don't need to pay a person at the counter to get the information. Also touchscreen can be equally accessible to everyone."} 

We also asked participants to provide feedback on the physical design of the Toucha11y bot. P2, P4, P5, and P6 stated that the device was easy to hold, had a good size, and would fit in their handbags or backpacks. However, for those who do not own a bag, the current prototype may be too large. P1 pointed out that the device was too large to fit in their bag or pocket, and if a user did not have a bag, the device was unlikely to be carried. P3 also expressed a desire for the size to be reduced so that they could be \textit{"put in the pocket and carry it out"} like a phone. 

Finally, the Toucha11y prototype also sparked discussions about the responsibility of technological accessibility. For example, P5 pointed out that the success of Toucha11y might have \textit{"unintended consequences"}---that technology companies who build touchscreen kiosks \textit{"would care even less (about accessibility) if they see this robot is so cool and using the phone and your app is not difficult at all." "Why do companies need to make technologies accessible?"}

\section{Discussion and Future Work}

\subsection{Improve Toucha11y Design}

\subsubsection{Reduce the size}
As a working prototype, the Toucha11y bot measures \SI{50}{\milli\metre} by \SI{70}{\milli\metre} by \SI{93}{\milli\metre}. As P1 and P3 pointed out, the current prototype can be too big to be carried around as a personal device. Its size also precludes its use on smaller touchscreen interfaces, like those found on treadmills. We are considering several optimizations for the next iteration of the Toucha11y bot. For example, the Pi Zero W, which measures \SI{65}{\milli\metre} by \SI{30}{\milli\metre}, is the single largest component in the current bot design. A VoCore 2.0 controller~\cite{vocore} with roughly one-third the size (\SI{25.6}{\milli\metre\squared}) but similar performance can be used as an alternative.

\begin{figure}[ht!]
  \includegraphics[width=\columnwidth]{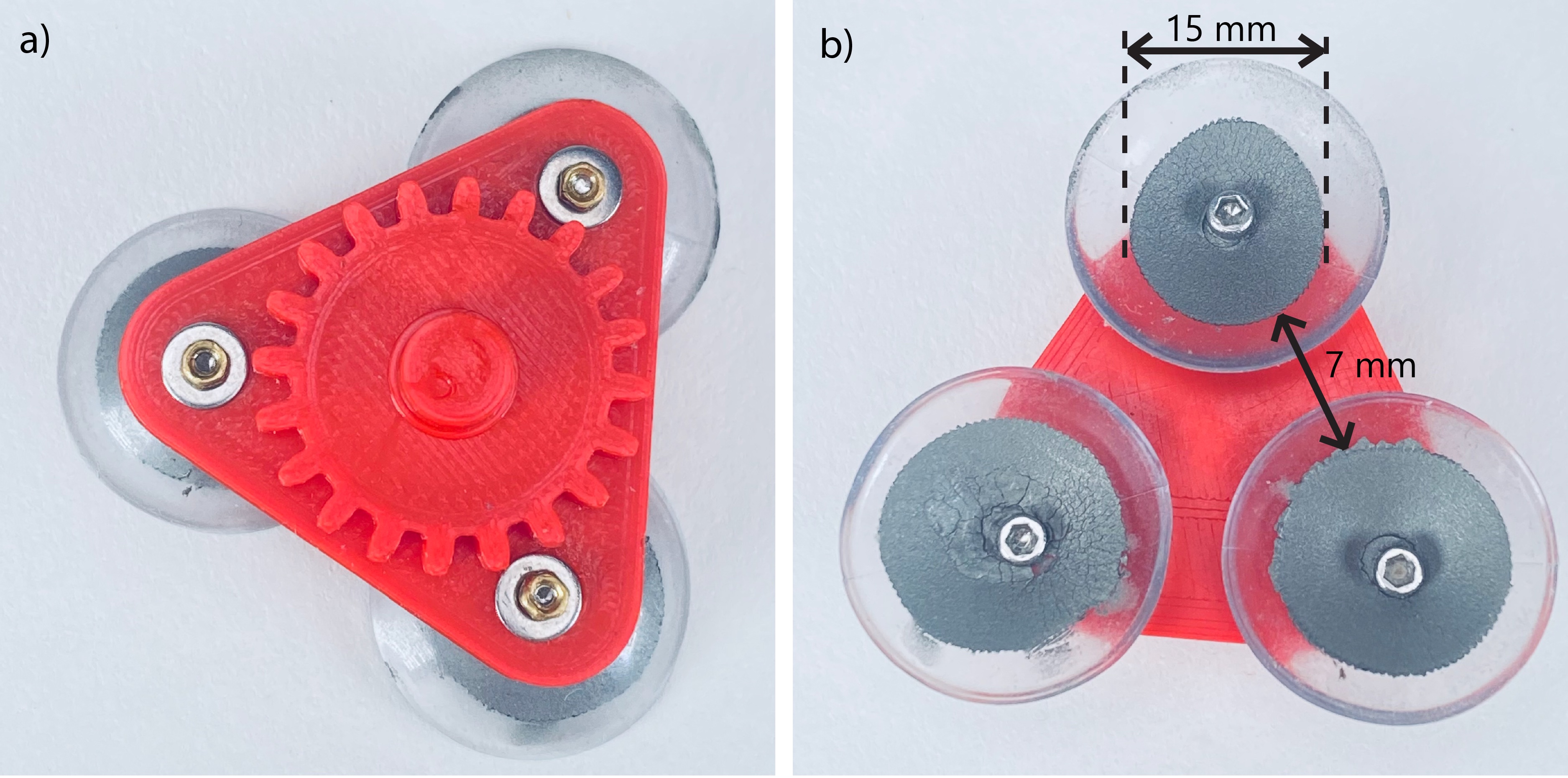}
  \caption{Suction cups with a conductive coating inside to trigger the touchscreen area below.}
  \Description{Suction cups with a conductive coating inside to trigger the touchscreen area below.}
  \label{fig:new base}
\end{figure}

\subsubsection{Cover touchable areas without relocation}

As the Toucha11y bot has to be placed directly on top of the touchscreen surface, the bot base will inevitably cover a portion of the screen area, which sometimes can be the place that needs to be activated.
In our current implementation, blind users must move the robot to a different location on the screen before proceeding.
However, this solution may introduce additional errors during relocation.

To reduce the need for relocation, we are currently experimenting with a new bot base design in which the suction cups can directly trigger touch events. 
As shown in Figure~\ref{fig:new base}, the inner surface of the base's suction cups (\SI{15}{\milli\metre} diameter) are coated with conductive paint, with a \SI{7}{\milli\metre} gap in-between. Since the gap distance is less than the recommended side length of a touchscreen button~\cite{thumb_size}, if the base is placed above an area to be activated, at least one suction cup will partially cover the button underneath and trigger accordingly.  

We plan to evaluate the reliability of the new base design and eventually incorporate it into a future version of the Toucha11y system.

\subsubsection{Improve privacy}
Using Toucha11y can partially alleviate privacy concerns, as blind users can enter sensitive information directly from their smartphone rather than seek help from strangers. 
However, it is still possible for those standing behind the user to see the bot's actions as it registers touch events. 
A potential solution is to redesign the bot's camera so that, in addition to taking photos of the kiosk touchscreen, it can also scan the user's surroundings to detect overlooking.
It is also possible to add a voice prompt to the smartphone interface to alert blind users.

\subsubsection{Avoid pre-labeled interfaces in the database}
Toucha11y assumes that the user interface of a touchscreen kiosk is pre-labeled and stored in a cloud-based database. The bot's camera is only to detect its placement, not to recognize the interface content. One limitation of this setup is that the bot cannot be used on a kiosk if its interface has not been pre-labeled or updated in the database. 

One future direction would be to have the Toucha11y bot's built-in camera capture the entire screen area and leverage potential modern computer vision algorithms (such as Tesseract 4~\cite{tesseract}) for screen interface recognition. For this to work, the camera height must be adjustable, i.e., it must be able to rise above the touchscreen high enough to capture the full screen.
While possible, enabling the height-changing camera will likely increase the bot's size and complexity and thus require further investigation.

\subsection{Long-Term Deployment}
Although the user evaluation results were promising, Toucha11y was only tested in one simulated environment for a limited amount of time. We intend to deploy the device for long-term use in order to better understand how blind users will use it in real life. For long-term deployment, in addition to the hardware improvements already mentioned, we also need to update the system software. For example, we will need to gather a much larger database of the labeled touchscreen interface to cover a wide range of actual in-use touchscreen user interfaces. We anticipate that similar crowdsourcing approaches~\cite{guo2019statelens,guo2017facade} can be integrated with Toucha11y to enlarge the interface database.

\subsection{Who is Responsible for Accessible Technology}
Toucha11y is a temporary solution to make existing, inaccessible touchscreen devices accessible to blind users. We strongly concur with P5's remark, as we do not want the research on Toucha11y to be the excuse for technology companies to stop developing accessible kiosks. Instead, we wish more technology businesses would begin updating public kiosks with accessibility features (e.g., ~\cite{mcdonald}) while also contributing the software interfaces of those inaccessible to the community so that they can be verified, labeled, and used by Toucha11y as a stopgap accessible solution.

\subsection{Beyond Touchscreen Accessibility}
Finally, while Toucha11y is designed to improve the accessibility of public kiosks for blind users, the notion of a small, personal device capable of mechanically interacting with the physical environment may present new opportunities for accessibility research. Looking forward, a Toucha11y-like bot could potentially be used to assist any people who have difficulty interacting with touchscreens, for example, people with motor disabilities. In the event that some touchscreen kiosks are poorly positioned and out of users' reach, a small mechanical bot with a long extendable reel can assist users in completing the touch interaction. 
If the bot's end-effector can be modified further, it could also be used to interact with physical gadgets beyond digital displays. For example, a Toucha11y-like bot with a universal gripper~\cite{amend2012positive} may serve as a tabletop assistant. After scanning the desk, it can physically grab objects of interest, e.g., a pill bottle, for blind users. We note, however, that each of the potential opportunities calls for extra examination and deep involvement from all target users.

\section{Conclusion}
We have presented Toucha11y, a technical solution to enable blind users to use existing inaccessible touchscreen kiosks independently and with minimal effort. 
Toucha11y consists of a mechanical bot, a mobile interface, and a back-end server. The bot can be instrumented to an arbitrary touchscreen of a kiosk by the blind user to recognize its content, retrieve the corresponding information from a database, and render it on the user's smartphone. Through the smartphone app, a blind user can access the touchscreen content, and make selections using the built-in accessibility features of a smartphone.
The bot can detect and activate the corresponding virtual button on the touchscreen. 
We presented the system design and a series of technical evaluations of Toucha11y. 
Through user evaluations, we concluded that Toucha11y could help blind users operate inaccessible touchscreen kiosks.

\begin{acks}
This work was supported in part by the New Direction Fund from the Maryland Catalyst Fund and a grant from the National Institute on Disability, Independent Living, and Rehabilitation Research (NIDILRR grant number 90REGE0008). NIDILRR is a Center within the Administration for Community Living (ACL), Department of Health and Human Services (HHS). This work does not necessarily represent the policy of NIDILRR, ACL, or HHS, and you should not assume endorsement by the Federal Government.
\end{acks}

\bibliographystyle{ACM-Reference-Format}
\bibliography{sample-base}

\appendix

\end{document}